\documentclass{article}

    \usepackage[nonatbib,preprint]{neurips_2020}

\usepackage[utf8]{inputenc} 
\usepackage[T1]{fontenc}    
\usepackage{hyperref}       
\usepackage{url}            
\usepackage{booktabs}       
\usepackage{amsfonts}       
\usepackage{nicefrac}       
\usepackage{microtype}      

\usepackage{graphicx}
\usepackage{amssymb,amsmath}
\usepackage{physics}
\usepackage{bm}
\usepackage{subcaption}

\usepackage{appendix}
\newcommand{\trm}[1]{\textrm{#1}}

\newcommand{\qqquad}{\qquad \qquad}

\newcommand{\T}{\intercal}

\title{Solving Differential Equations Using Neural Network Solution Bundles}

\author{%
   Cedric Flamant\\
   Department of Physics \\
   Harvard University \\
   Cambridge, MA, 02138 \\
   \texttt{cflamant@g.harvard.edu} \\
   \And
   Pavlos Protopapas\\
   John A. Paulson School of Engineering and Applied Science \\
   Harvard University \\
   Cambridge, MA, 02138 \\
   \texttt{pavlos@seas.harvard.edu} \\
   \And
  David Sondak \\
  John A. Paulson School of Engineering and Applied Science\\
  Harvard University\\
  Cambridge, MA 02138 \\
  \texttt{dsondak@seas.harvard.edu} \\
}

\begin{document}

\maketitle

\begin{abstract}
The time evolution of dynamical systems is frequently described by ordinary differential equations (ODEs), which must be solved for given initial conditions. Most standard approaches numerically integrate ODEs producing a single solution whose values are computed at discrete times. When many varied solutions with different initial conditions to the ODE are required, the computational cost can become significant. We propose that a neural network be used as a solution bundle, a collection of solutions to an ODE for various initial states and system parameters. The neural network solution bundle is trained with an unsupervised loss that does not require any prior knowledge of the sought solutions, and the resulting object is differentiable in initial conditions and system parameters. The solution bundle exhibits fast, parallelizable evaluation of the system state, facilitating the use of Bayesian inference for parameter estimation in real dynamical systems.
\end{abstract}

\section{Introduction}
Many dynamical systems are described by ordinary differential equations (ODEs) which relate the rates and values of state variables and external driving functions. While some simple ODEs have closed form solutions to them, the vast majority have to be solved approximately using discretization of the domain or by using spectral methods\cite{bernardi_spectral_1997}. The former approximating methods are more general, with Runge-Kutta and multi-step methods as typical examples. These methods seek to numerically integrate the ODEs, starting from initial conditions and stepping forward until the desired final time is attained. While these conventional methods are generally efficient for determining the state of a system for a sequence of times, if we are only interested in the state at a specific later time, substantial computational effort must still be expended determining all the states at steps leading up to the state of interest. This causal order also limits parallelizability of the conventional single- and multi-step methods since the task cannot be parallelized in time---until the preceeding state is known, processors tasked with finding a segment of the system's evolution over a later time interval cannot start calculating the correct piece of the trajectory.

\textbf{Previous work}
\nopagebreak

The idea of approximating the solutions of differential equations with neural networks was first developed by Dissanayake and Phan-Thien, where training was accomplished by minimizing a loss based on the network's satisfaction of the boundary conditions and the differential equations themselves\cite{dissanayake_neural-network-based_1994}. Lagaris \textit{et al.} showed that the form of the network could be chosen to satisfy boundary conditions by construction, and that automatic differentiation could be used to compute the derivatives appearing in the loss function\cite{lagaris_artificial_1998}. After training, the approximate value of the solution at any point within the training range can be computed in constant time without having to compute previous states first. This method has been extended to systems with irregular boundaries\cite{lagaris_neural-network_2000,mcfall_artificial_2009}, applied to solving PDEs arising in fluid mechanics\cite{baymani_artificial_2010}, and software packages have been developed to facilitate its application\cite{lu_deepxde_2020,koryagin_pydens_2019,chen_neurodiffeq_2020}. In the Lagaris approach, the neural network learns a single solution to the ODE. For different sets of initial conditions, or for different sets of parameters in the differential equation, the network has to be retrained on the new task.

\textbf{Our contributions}

We propose an extension of the Lagaris method\cite{lagaris_artificial_1998} where the neural network is instead taught a variety of solutions to a parameterized differential equation. This increases the reusability of the trained network and can speed up tasks that require knowing many solutions, such as for Bayesian parameter inference, propagating uncertainty distributions in dynamical systems, or inverse problems. While it is straightforward to extend our approach to all the situations considered in the Lagaris paper, \textit{i.e.} for problems containing various types of boundary conditions, for partial differential equations and higher derivatives, we focus on initial value problems in first-order ODEs. We show that our method has promise when applied to a variety of tasks requiring quick, parallel evaluation of multiple solutions to an ODE, and where it is useful to be able to differentiate the state at a particular time with respect to the initial conditions or ODE parameters. Our contributions also include weighting the loss function to improve global error and demonstrating an application of curriculum learning to help with training. With the rapid advances in neural network development, as well as its supporting hardware, employing this method will become cheaper and more efficient, further extending its applicability.

\section{Solution Bundles}

When working with ODEs it is common to require multiple solutions corresponding to different initial conditions. In dynamical systems, each of these solutions represents a trajectory, tracing out an alternate time evolution of the system state. In addition, when an ODE is parameterized, say by a physical constant whose value has an associated uncertainty, it can be useful to have different solutions for various values of the parameters.


\subsection{Method Description}

Consider the following general first-order differential equation parameterized by $\bm{\theta}$:
\begin{align}
  \vb{G}\qty(\vb{x}, \dv{\vb{x}}{t}, t \; ; \: \bm{\theta}) = 0,
  \label{eq:lagarisextended}
\end{align}
where $\vb{x} \in \mathbb{R}^n$ is a vector of state variables, $t$ is time, and $\bm{\theta} \in \mathbb{R}^p$ are physical parameters associated with the dynamics. We assume that the ODE describes a deterministic system where initial conditions $\vb{x}_0$ of the state variables uniquely determines a solution. The solutions to Eq. (\ref{eq:lagarisextended}) over the time range $[t_0, t_f]$, together with a subset of initial conditions $X_0 \subset \mathbb{R}^n$, and a set of parameters $\Theta \subset \mathbb{R}^p$, define a \emph{solution bundle} $\vb{x}(t; \vb{x}_0, \bm{\theta})$, where $\vb{x_0}\in X_0$, $\bm{\theta} \in \Theta$.

Let the approximating function to Eq. (\ref{eq:lagarisextended}) for the solution bundle over $(X_0, \Theta)$ be given by 
\begin{align}
  \hat{\vb{x}}(t \: ;\: \vb{x}_0, \bm{\theta}) = \vb{x}_0 + a(t)\vb{N}(t \: ;\:\vb{x}_0, \bm{\theta} \: ; \: \vb{w}),
  \label{eq:approximatingfunction}
\end{align} 
where $\vb{N} : \mathbb{R}^{1+n+p} \to \mathbb{R}^n$ is a neural network with weights $\vb{w}$, and $a : [t_0,t_f] \to \mathbb{R}$ satisfies $a(t_0) = 0$. This form explicitly constrains the trial solution to satisfy the initial condition $\vb{x}(t_0) = \vb{x}_0$. The choice of $a(t)$ can affect the ease of training the network. While $a(t) = t - t_0$ is sufficient, Mattheakis \textit{et al.} demonstrated that $a(t) = 1 - e^{-(t-t_0)}$ results in better convergence in the Lagaris method\cite{mattheakis_hamiltonian_2020}, and we observe similar benefits in our extension, likely due to its diminishing effect farther from the initial time. We primarily use multilayer fully-connected neural networks in our experiments, but it is worth exploring other architectures in the future.

The unsupervised loss function used in training has the form 
\begin{align}
  L = \frac{1}{\abs{B}} \sum_{(t_i,\vb{x}_{0i}, \bm{\theta}_i) \in B} b(t_i)\abs{\vb{G}\qty(\hat{\vb{x}}(t_i \: ; \: \vb{x}_{0i}, \bm{\theta}_i), \pdv{\hat{\vb{x}}(t_i \: ; \: \vb{x}_{0i}, \bm{\theta}_i)}{t}, t_i \: ; \: \bm{\theta}_i)}^2,
  \label{eq:generalloss}
\end{align}
where the set $B = \qty{(t_i,\vb{x}_{0i}, \bm{\theta}_i)}$ constitutes a training batch of size $\abs{B}$, with $t_i \in [t_0,t_f]$, $\vb{x}_{0i} \in X_0$, and $\bm{\theta}_i \in \Theta$ drawn from some distribution over their respective spaces. The function $b : [t_0,t_f] \to \mathbb{R}$ appearing in Eq. (\ref{eq:generalloss}) is used to weight data points based on their time, as will be discussed later. We typically use $b(t) = \exp\qty(-\lambda (t-t_0))$, with hyperparameter $\lambda >0$. We found that uniform sampling over the spaces $[t_0,t_f]$, $X_0$, and $\Theta$, usually works well in practice, but there are situations where it is helpful to use a different distribution to generate batches. For example, for the FitzHugh-Nagumo model discussed in Section \ref{section:fhnmodel}, a batch-number-dependent distribution over times $[t_0,t_f]$ was used for curriculum learning. Finally, in the loss function, the time derivative of $\hat{\vb{x}}(t)$ is computed using automatic differentiation.

There is no concept of an epoch in the training process of this method since every batch will be a unique sample of size $\abs{B}$ from the distribution across times, initial conditions, and parameters. This is similar to meshfree implementations of the Lagaris method, which have been shown to work better than operating on a fixed grid\cite{sirignano_dgm_2018}. The model cannot overfit as we are effectively operating in a regime of infinite data. The training ends when the approximation to the solution bundle is deemed acceptable, based on either the history of past losses or on some other metric, like its difference compared to a few solution curves computed via conventional methods. Ideally, the training process should be carried out until the network weights are locally optimal, though in general the approximation will still differ from the exact solution. To obtain incrementally better solution bundles, one could increase the complexity of the network since the universal approximation theorem\cite{cybenko_approximation_1989,sonoda_neural_2017} guarantees the existence of a better approximating network to the true solution bundle.


At the end of training, the neural network solution bundle $\hat{\vb{x}}(t; \vb{x}_0, \bm{\theta})$ can be used to evaluate, in constant execution time, the approximate value of state $\vb{x}$ at any time $t \in [t_0, t_f]$ for any initial condition $\vb{x}_0 \in X_0$ and parameters $\bm{\theta} \in \Theta$. Our approach is highly amenable to parallelization both during the training and inference stages. The training is parallelized in time by distributing the points in batch $B$ across each processing unit, as seen in Section \ref{subsubsection:trainingsol}. During the inference stage, the evaluation of the neural network can be parallelized, unlike in stepping methods where steps have to be computed sequentially, and each step consists of too few operations to benefit from parallelization. These strengths of our method are useful when the behavior of a distribution of solutions is desired, such as for propagating state uncertainty or for performing Bayesian inference.

The neural network solution bundle also has a closed analytic form and is differentiable in all of its inputs. This capability can be used for a variety of useful tasks. For example, in Bayesian estimation differentiability in the initial conditions and ODE parameters enables gradient-based maximization, simplifying the calculation of maximum a posteriori (MAP) estimates of these quantities. It also simplifies the application of ``shooting methods,'' where a condition at a later time is known and the parameters and initial conditions that are consistent with the constraint are sought. Differentiability can also be useful for more general optimization tasks and sensitivity studies.


\subsubsection{Discussion of the Weighting Function.}

Given that the approximating function Eq. (\ref{eq:approximatingfunction}) will generally not be able to perfectly satisify the target ODE everywhere, there will always be some bias error or epistemic uncertainty regardless of the choice of $\vb{w}$. If we do not weight the loss function Eq. (\ref{eq:generalloss}), \textit{i.e.} $b(t) = 1$, we do not get to influence how the local error
\begin{align}
  \bm{\varepsilon}(t_i \: ; \: \vb{x}_{0i}, \bm{\theta}_i) \equiv \vb{G}\qty(\hat{\vb{x}}(t_i \: ; \: \vb{x}_{0i}, \bm{\theta}_i), \pdv{\hat{\vb{x}}(t_i \: ; \: \vb{x}_{0i}, \bm{\theta}_i)}{t}, t_i \: ; \: \bm{\theta}_i)
  \label{eq:localerror}
\end{align}
is distributed across the training region. However, by applying an exponentially decaying weight $b(t) = e^{-\lambda (t - t_0)}$, we can convey in the loss function the larger contribution of early-time local errors to our metric of interest, the global error.
An appeal for using a decaying exponential weighting comes from considering an upper-bound on the global error. Consider a differential equation of the form $\dd{x}/\dd{t} = f(t,x)$. The local error Eq. (\ref{eq:localerror}) is $\varepsilon(t) = \dd{\hat{x}}/\dd{t} - f(t,\hat{x})$, and it can be shown (Appendix A) that the global error $\epsilon(t)=\hat{x}(t) - x(t)$ is bounded by
\begin{align}
  \abs{\epsilon(t)} \leq \frac{\varepsilon_{t'}}{L_f} \qty(e^{L_f \qty(t - t_0)} - 1),
  \label{eq:globalerror}
\end{align}
where $L_f$ is the Lipschitz constant of $f$, and $\varepsilon_{t'} \equiv \max_{t_0 \leq t \leq t'} \abs{\varepsilon(t)}$. This bound shows that global error can grow exponentially as a result of an early local error, so an exponential form for the weighting function $b(t)$ incorporates this relative importance in the loss. 

Note that in the above discussion we assume scarce knowledge about the specifics of an ODE, showing that an exponential weighting is generally sensible. Specific weighting functions can likely be tailored to outperform exponential weighting for a given ODE.

\section{Propagating a Distribution}

A neural network solution bundle provides a mapping from initial states to states at later times. This can be useful for time-evolving a distribution over initial conditions to obtain a probability distribution over states at later time $t$. Given a probability density over initial states $p_{0}(\vb{x}_0)$, we note that the solution bundle $\vb{x}(t \: ; \: \vb{x}_0)$ at time $t$ describes a coordinate transformation from $\vb{x}_0$ to $\vb{x}_t$. In practice we can simply sample the initial state space and construct a histogram of output $\vb{x}_t$ states using the solution bundle. The sampling of initial states can be done according to $p_0\qty(\vb{x}_0)$ using Markov chain Monte Carlo (MCMC) methods, or if the dimensionality of the state vector is low enough, by simply performing a uniform sampling over the initial states and constructing a weighted histogram of $\vb{x}_t$, weighting each sample by $p_0\qty(\vb{x}_0)$ for the $\vb{x}_0$ that generated it. Appendix B explains how an analytic expression for the transformed probability density can also be obtained by making use of the differentiability of the solution bundle.

\subsection{Planar Circular Restricted Three-Body Problem}

The planar circular restricted three-body problem describes a special case of the motion of three masses under Newton's law of universal gravitation. This special case describes the motion of the third body, which is assumed to have negligible mass, in the co-rotating frame of the first two bodies in circular orbits around their barycenter. All three bodies are also assumed to lie in the same plane ($x$-$y$ plane), with no velocity in the $z$ direction. For clarity of discussion, let body 1 be the Earth, body 2 be the Moon, and body 3 be an asteroid. The asteroid has position $\vb{r}(t) = \qty(x(t),y(t))^\T$ and velocity $\vb{u}(t) = \qty(u(t), v(t))^\T$. We will call the full state vector $\vb{q} = \qty( \vb{r}^\T, \vb{u}^\T)^\T$. The nondimensionalized mass of the Earth is given by $m_1 = 1 - \mu$, and the mass of the Moon is $m_2 = \mu$, where $\mu$ is the ratio of the mass of the the Moon to the total mass of the pair. The nondimensionalized equations of motion of the asteroid are given by\cite{szebehely_theory_1967}:
\begin{align}
  &\dv{x}{t} = u\;, \qquad \dv{u}{t} = x - \mu + 2v - \qty[ \frac{\mu(x-1)}{\qty\big( \qty(x-1)^2 + y^2)^{3/2}} + \frac{(1 - \mu) x}{\qty(x^2 + y^2)^{3/2}}],& \\
  &\dv{y}{t} = v\;, \qquad \dv{v}{t} = y - 2u - \qty[  \frac{\mu y}{\qty\big( \qty(x-1)^2 + y^2)^{3/2}} + \frac{\qty(1 - \mu) y}{ \qty(x^2 +y^2)^{3/2}}].&
  \label{eq:crtbEOM}
\end{align}

\subsubsection{Training the Solution Bundle}
\label{subsubsection:trainingsol}

We trained a fully-connected MLP with 8 hidden layers of 128 neurons per layer uniformly over the $x_0$, $y_0$, $u_0$, $v_0$ initial condition space $X_0 = [1.05,1.052] \times [0.099, 0.101] \times [-0.5, -0.4] \times [-0.3, -0.2]$ and times $[-0.01, 5]$, and used a fixed parameter $\mu = 0.01$. Even though we only intend to evaluate the solution bundle at times $t \in [0,5]$, we found that including times slightly earlier than $t_0$ in training helps improve accuracy. This makes the approximating function satisfy the ODE on both sides of $t_0$, resulting in a more accurate value of the derivative term in Eq. (\ref{eq:generalloss}) around $t = t_0$. We used batchsize $\abs{B} = 10,000$, the Adam optimizer\cite{kingma_adam_2017}, and learning rate $\eta = 0.001$, which we reduced on plateau. For the weighting function $b(t)$ in the loss, Eq. (\ref{eq:generalloss}), we chose $b(t) = \exp\qty(-\lambda t)$ where $\lambda = 2$.

\begin{figure}[h!]
  \centering 
  \includegraphics[width=0.24\textwidth]{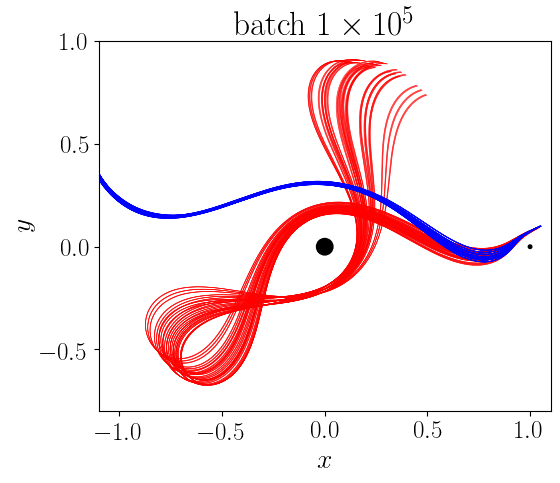}
  \includegraphics[width=0.24\textwidth]{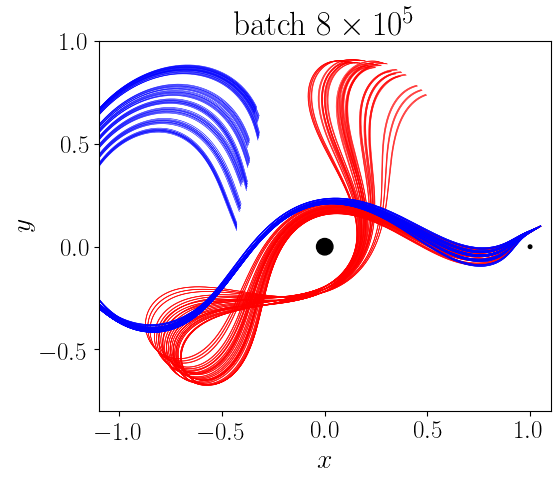}
  \includegraphics[width=0.24\textwidth]{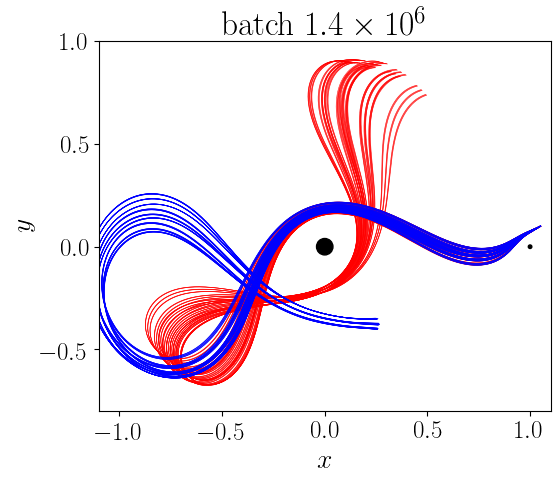}
  \includegraphics[width=0.24\textwidth]{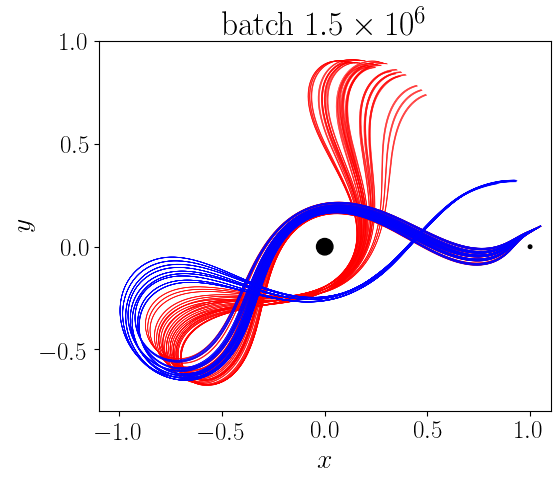}
  \includegraphics[width=0.24\textwidth]{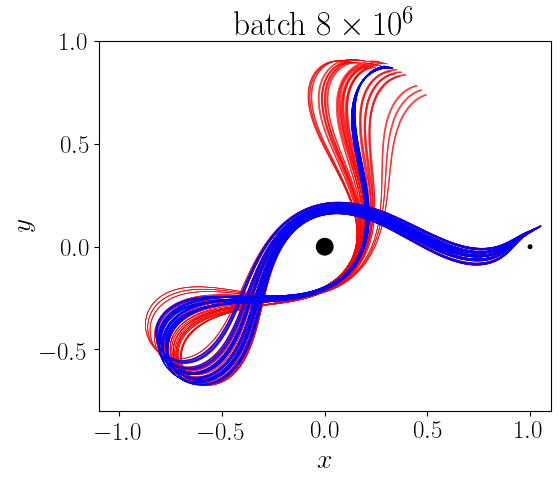}
  \includegraphics[width=0.24\textwidth]{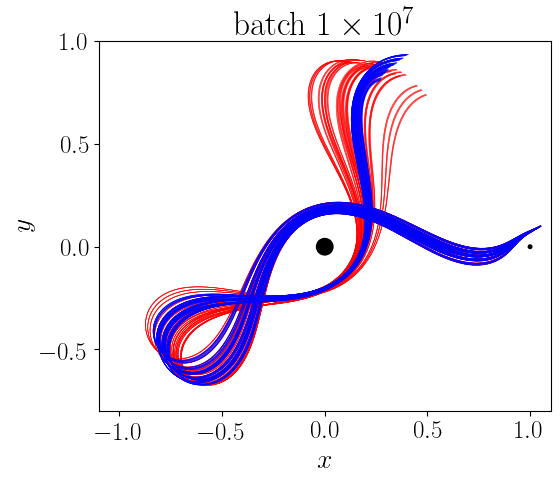}
  \includegraphics[width=0.24\textwidth]{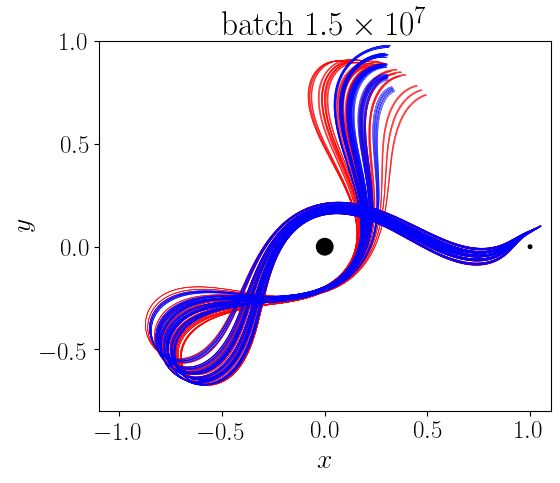}
  \includegraphics[width=0.24\textwidth]{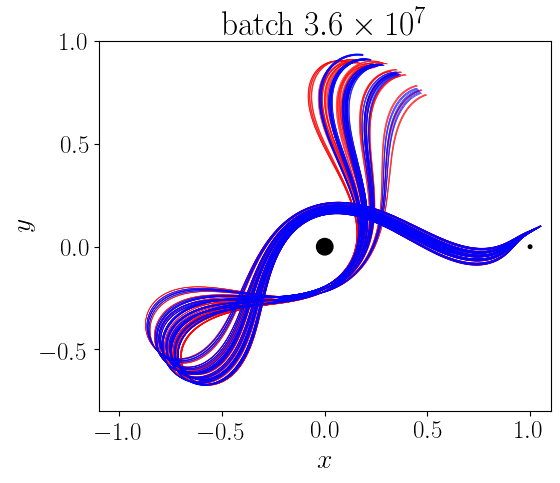}
  \caption{Plots of a few trajectories from the neural network solution bundle at various points in the training. Red trajectories are calculated with fourth-order Runge-Kutta, and the neural network solutions are shown in blue.}
  \label{fig:earthmoontrain}
\end{figure}

Figure \ref{fig:earthmoontrain} shows the adaptation of the neural network solution bundle during training with increasing batch number. It also demonstrates the parallelization in time of the method, with the curvature of later parts of the trajectory adjusting even before the earlier states have settled. This is reminiscent of the dominant approach for parallel-in-time solving of differential equations, Parareal\cite{lions_resolution_2001}, which also involves computation of approximate later trajectories before earlier path segments are precisely known.

\subsubsection{Propagation of Uncertainty}

With the neural network solution bundle trained, we can use it to propagate distributions in time. Suppose we have two measurements of the position of an asteroid at two different times, along with some uncertainty. If the majority of the probability mass of these uncertainty distributions falls within the solution bundle, it is straightforward to compute a probability distribution for the future position of the asteroid by constructing a histogram weighted by the uncertainty distribution of the observations. Details of the calculation can be found in Appendix B.

\begin{figure}[h]
  \centering 
  \includegraphics[width=0.32\textwidth]{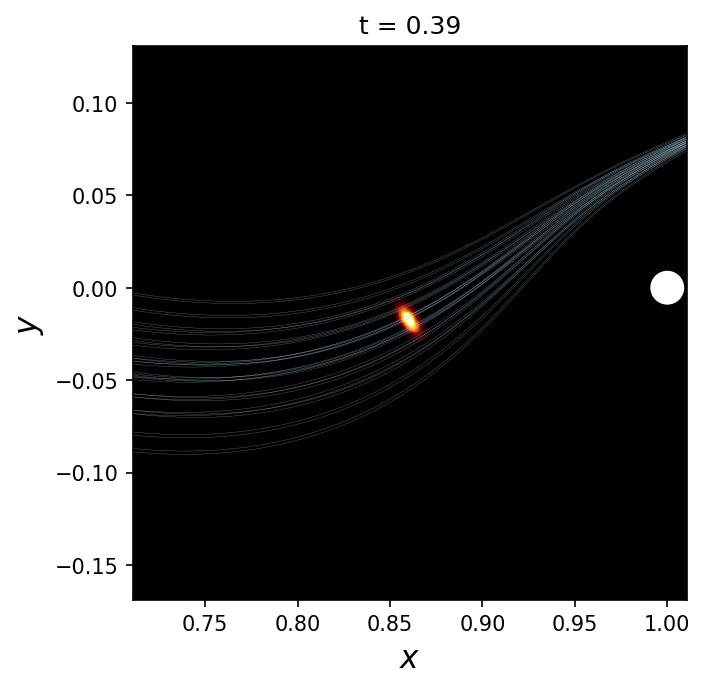}
  \includegraphics[width=0.32\textwidth]{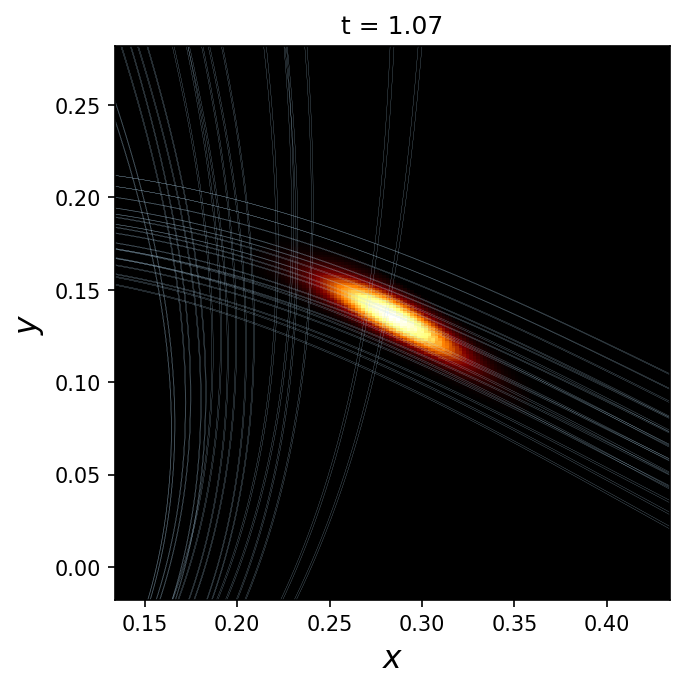}
  \includegraphics[width=0.32\textwidth]{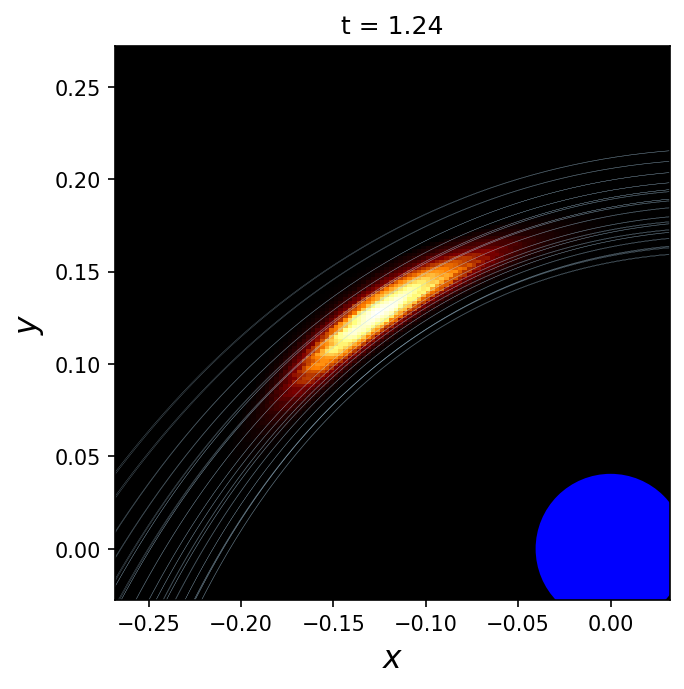}
  \caption{Probability distribution $p\qty(\vb{x}(t) = (x,y)^\T)$ at various times. A few trajectories in the bundle are shown in white, and the distribution is shown as a heatmap. The full path is shown in Figure \ref{fig:earthmoontrain} and the full trajectory of the distribution is in Appendix B.1.}
  \label{fig:earthmoondist}
\end{figure}

\section{Bayesian Parameter Inference}

Another common task that requires computing many solutions to an ODE is Bayesian parameter inference in systems described by differential equations. In the physical sciences and other fields employing mathematical modeling of data, it is often necessary to estimate parameters of a system based on experimental measurements, as well as to determine their uncertainties. If the system is described by differential equations, these parameters modify terms in the equations, resulting in different families of solutions. The probability density of the initial conditions and parameters $ \vb{x}_0, \bm{\theta}$ given a set of observed data $\qty{(t_i,\vb{x}_i)}$ and prior $p(\vb{x}_0, \bm{\theta})$ can be computed from Bayes' theorem,
\begin{align}
  p\qty\big(\vb{x}_0, \bm{\theta} \mid \qty{(t_i,\vb{x}_i)}) = \frac{p\qty\big(\qty{(t_i,\vb{x}_i)} \mid \vb{x}_0, \bm{\theta}) p\qty( \vb{x}_0, \bm{\theta})}{p(\qty{(t_i,\vb{x}_i)})} \propto p\qty\big(\qty{(t_i,\vb{x}_i)} \mid \vb{x}_0, \bm{\theta}) p\qty(\vb{x}_0,\bm{\theta}).
  \label{eq:bayesparam}
\end{align}
Determination of the likelihood $p\qty\big(\qty{(t_i,\vb{x}_i)} \mid \vb{x}_0, \bm{\theta})$ is typically the computationally intensive step as it requires computing $\qty{\qty\big(t_i, \vb{x}(t_i \: ;\:\vb{x}_0, \bm{\theta}))}$ for parameters $\vb{x}_0, \bm{\theta}$ to compare to the data $\qty{(t_i,\vb{x}_i)}$. Evaluating $\vb{x}(t_i \: ;\: \vb{x}_0, \bm{\theta})$ with conventional methods would require forward stepping from the initial conditions all the way to time $t_i$, and this process would have to be repeated for every different set of initial states and parameters $\vb{x}_0, \bm{\theta}$. The greater the desired precision of the posterior distribution for the parameters, the more often the differential equation has to be solved, which could be potentially millions of times. However, if a neural network solution bundle has been trained over $X_0$ and $\Theta$ containing the expected range of initial conditions and parameters, $\hat{\vb{x}}(t_i \: ; \:\vb{x}_0, \bm{\theta})$ can be calculated in constant time for any $t_i \in [t_0, t_f]$, and the entire set of points $\qty{\qty\big(t_i, \hat{\vb{x}}(t_i \: ;\: \vb{x}_0, \bm{\theta}))}$ can be computed in parallel. This allows for rapid likelihood evaluation and more efficient Bayesian inference. The training cost of the solution bundle can be further offset if it is used for many different sets of data. In effect, a trained neural network solution bundle for an often-used ODE could be shared amongst research groups, cutting back on the number of redundant calculations performed globally.

\subsection{Rebound Pendulum}

\begin{figure}[h!]
  \centering
  \includegraphics[trim={42bp 10bp 40bp 40bp},clip,width=0.3\textwidth]{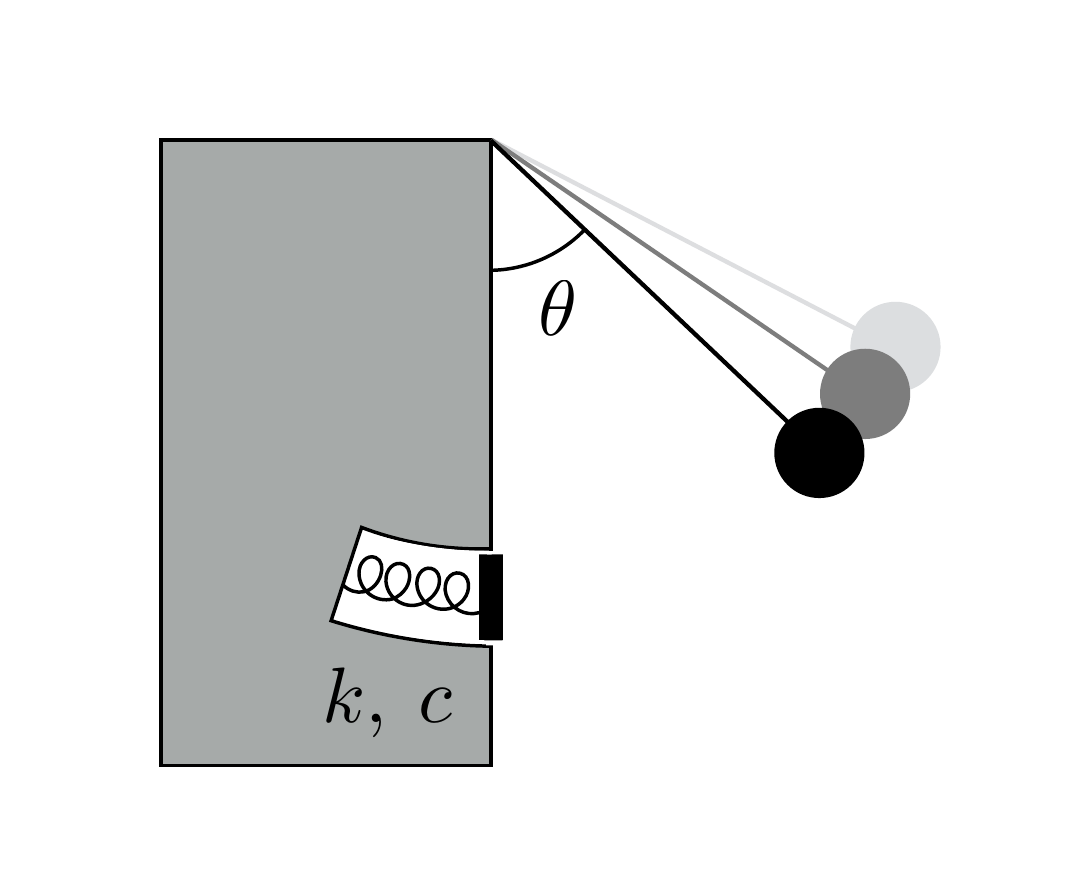}
  \includegraphics[width=0.30\textwidth]{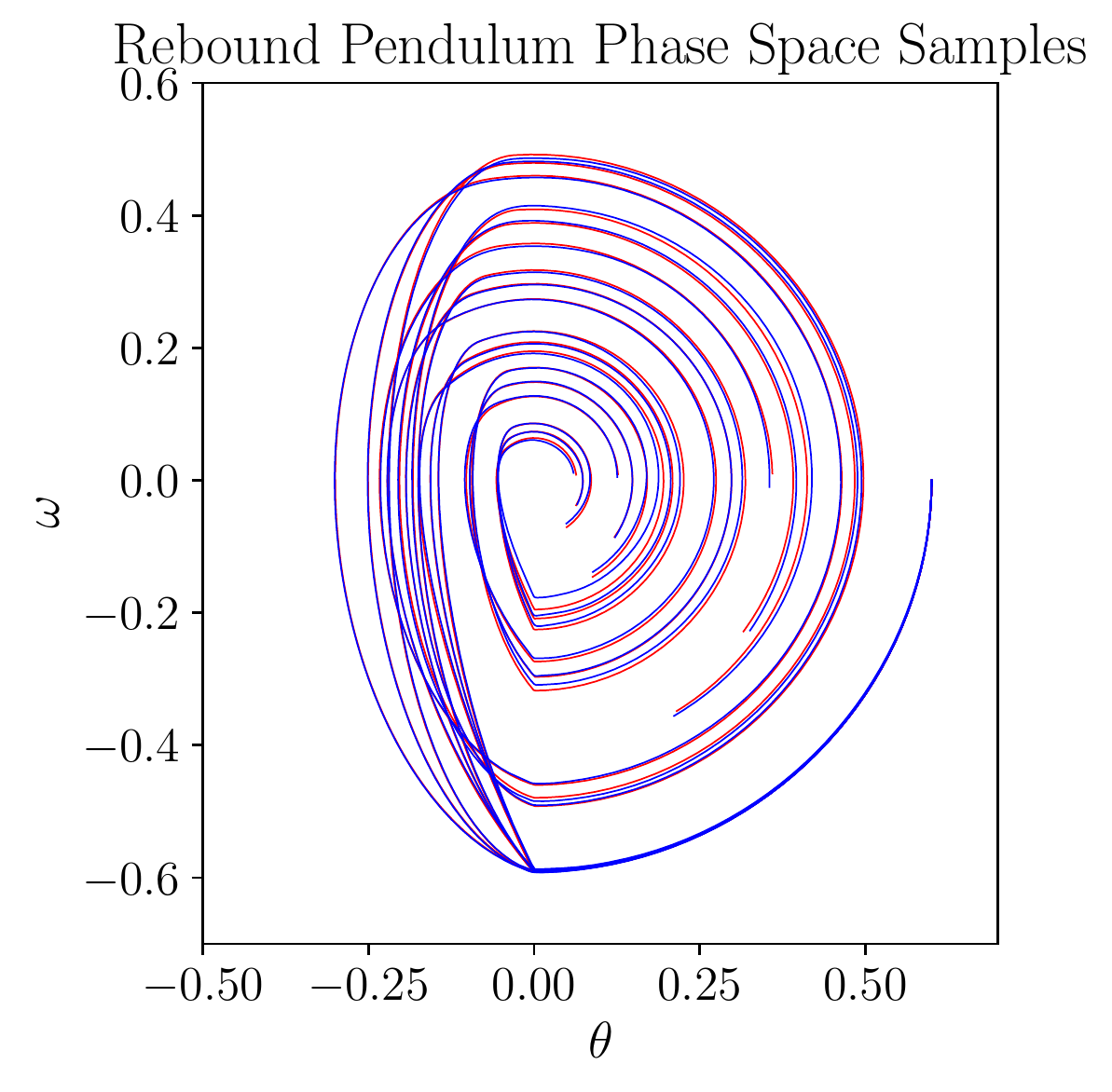}
  \includegraphics[width=0.37\textwidth]{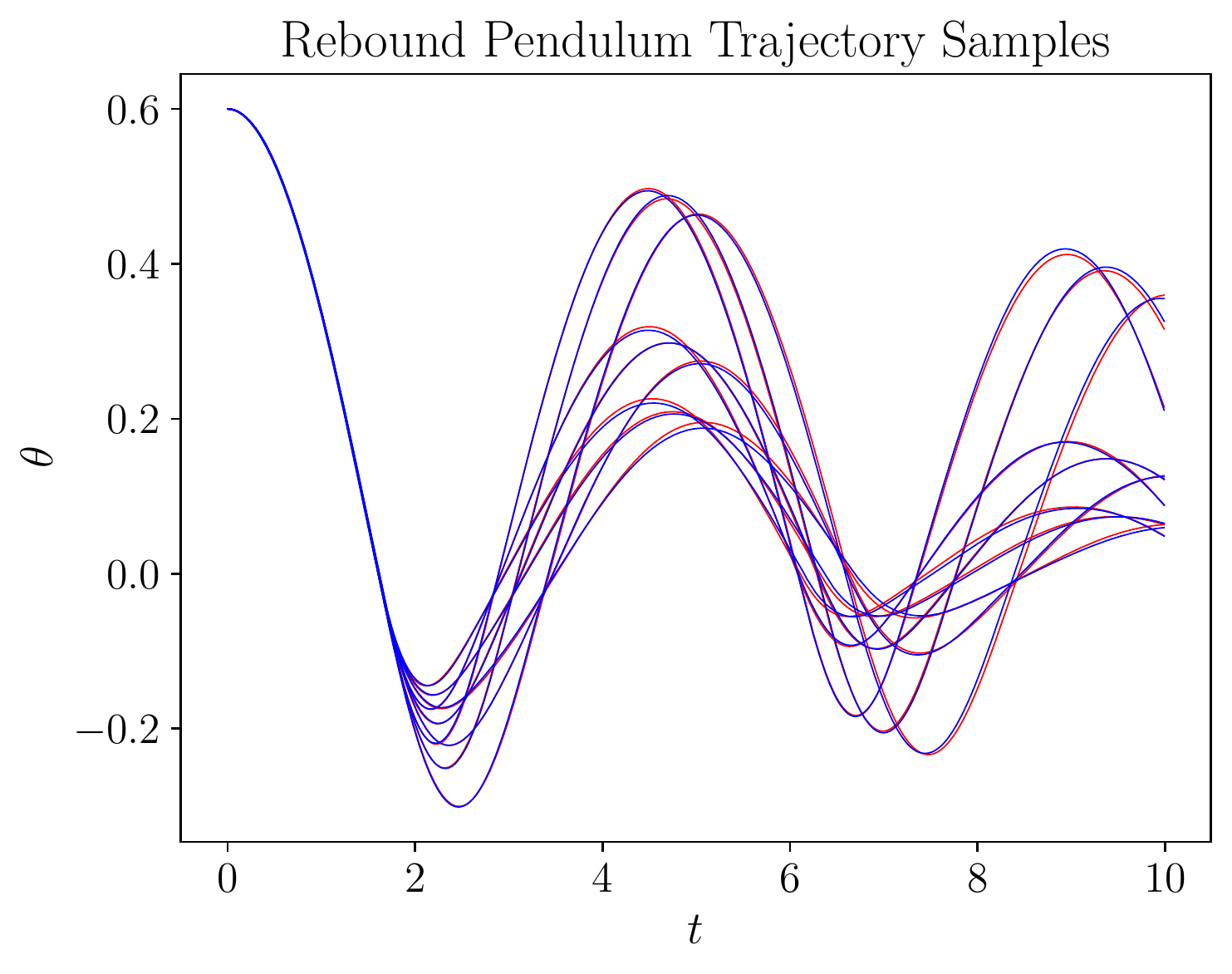}
  \caption{\textbf{Left:} Diagram of a rebound pendulum. \textbf{Center and Right:} A selection of solutions within the rebound pendulum solution bundle. The red reference curves are computed with Runge-Kutta and the trajectories from the solution bundle are overlayed in blue.}
  \label{fig:reboundpendulumtraj}
\end{figure}

\begin{figure}[h!]
  \centering 
  \includegraphics[width=0.43\textwidth]{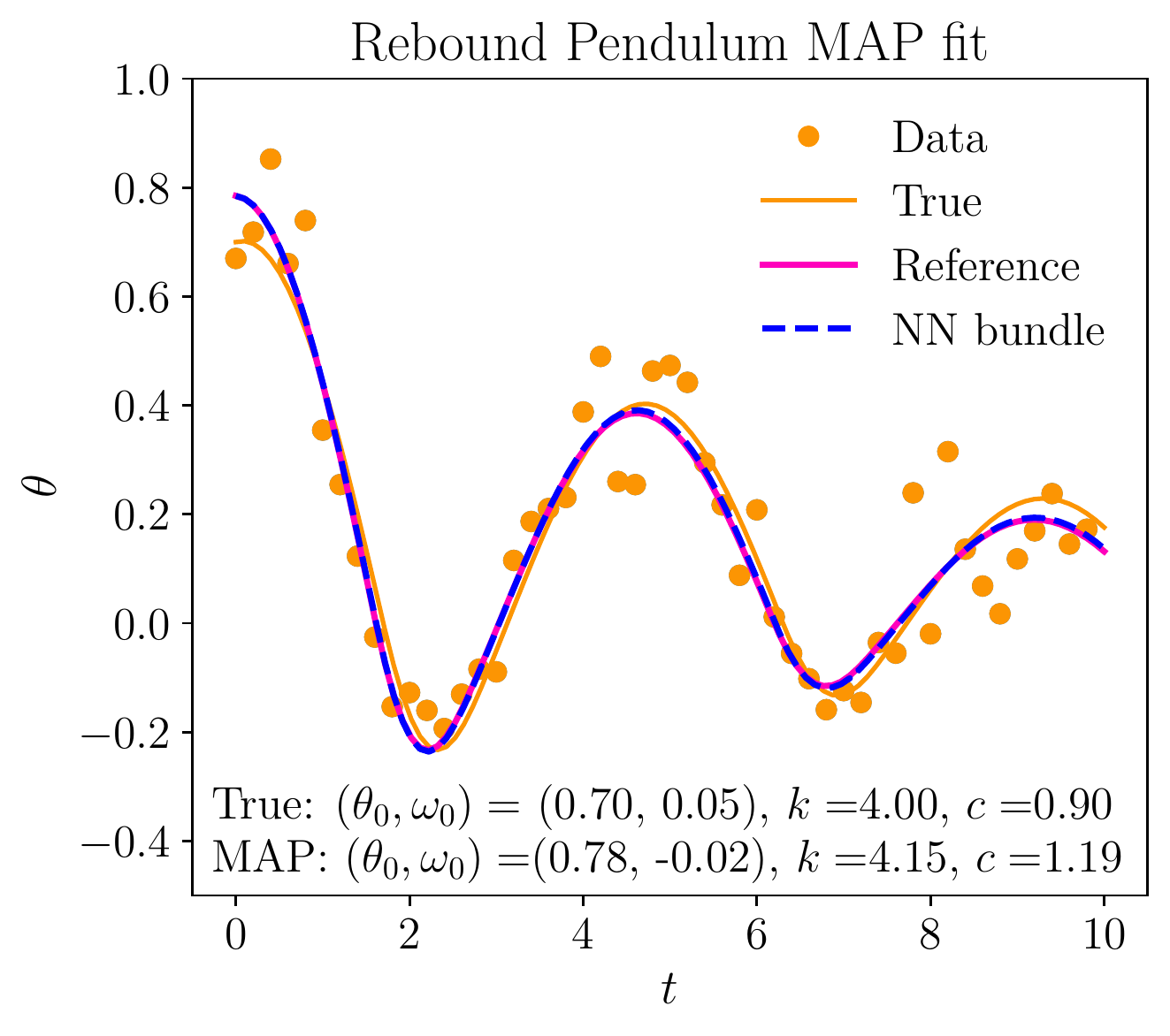}
  \includegraphics[width=0.56\textwidth]{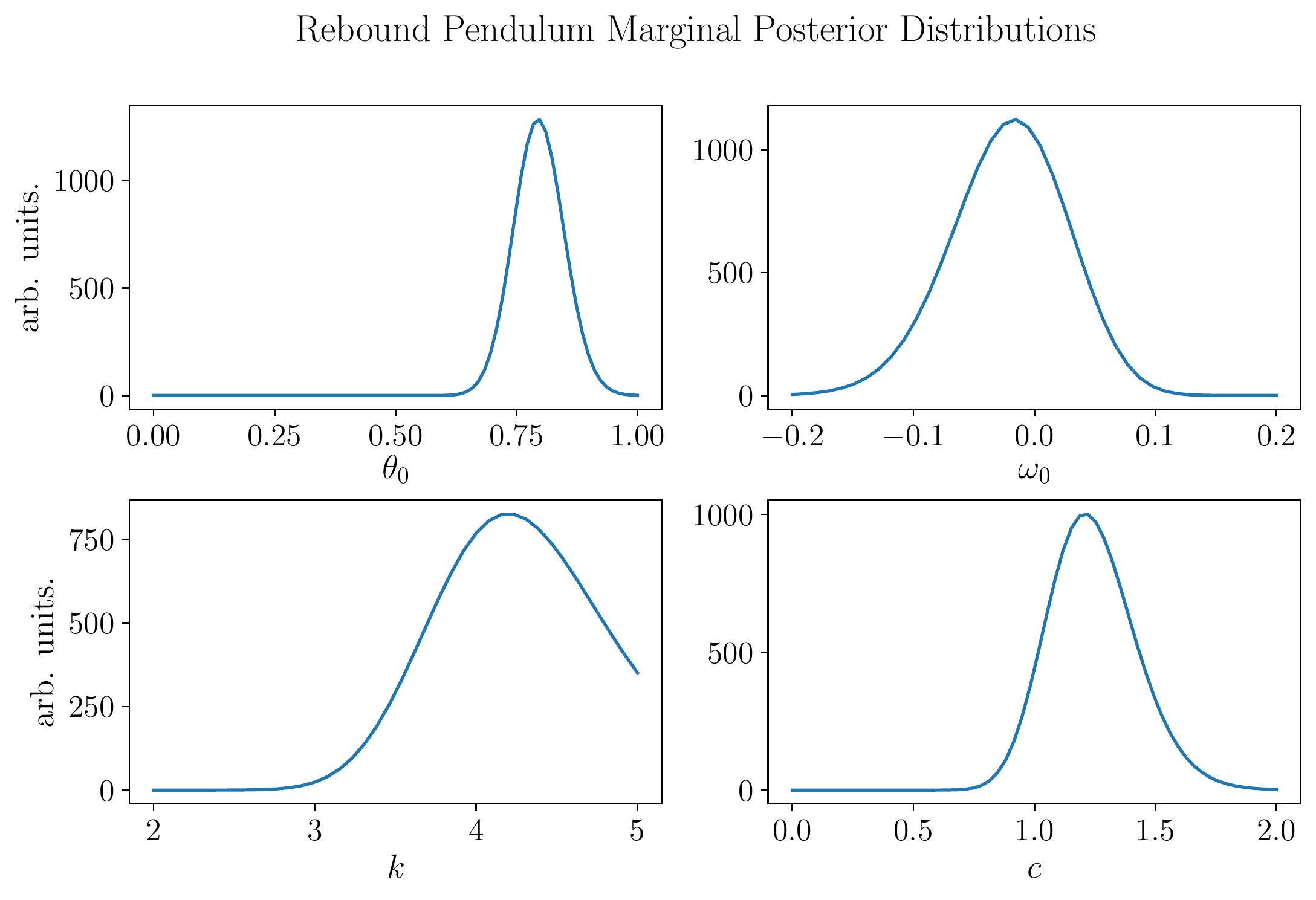}
  \caption{\textbf{Left:} Rebound pendulum fit corresponding to maximum a posteriori estimate. A Runge-Kutta reference curve using the MAP parameters is also plotted. \textbf{Right:} Unnormalized marginal posterior distributions of the unknown initial conditions and parameters.}
  \label{fig:reboundpendulumfit}
\end{figure}

A rebound pendulum consists of a simple pendulum that can collide with a damped spring at the bottom of its swing. A diagram of the setup is shown in Figure \ref{fig:reboundpendulumtraj}. The equations of motion for the state vector $\vb{x} = (\theta, \omega)^\T$ of the pendulum are given by
\begin{align}
  \dv{\theta}{t} = \omega, \qquad \dv{\omega}{t} = - \frac{g}{\ell} \sin \theta + H(-\theta) \trm{ReLU}\qty(-k \theta  - c \omega),
  \label{eq:reboundpendulumeomb}
\end{align}
where $\trm{ReLU}(x) = \max(x,0)$, $H(x)$ is the Heaviside step function, $g$ is the gravitational acceleration, $\ell$ is the length of the pendulum, $k$ is the spring constant, and $c$ is the damping coefficient. Note that here $\theta$ is the angle of the pendulum and not a parameter.
A few example solutions are shown in Figure \ref{fig:reboundpendulumtraj}, where the convergence of the neural network approximation can be compared to reference solutions computed with Runge-Kutta.

Using the trained solution bundle, we can fit simulated angle measurements which have Gaussian error. The result of Bayesian inference of the initial conditions and parameters is shown in Figure \ref{fig:reboundpendulumfit}, presenting both the MAP estimate and the marginal posterior distributions for a uniform prior.

\section{Curriculum Learning: FitzHugh-Nagumo Model}
\label{section:fhnmodel}

\begin{figure}[h!]
  \begin{subfigure}{.5\textwidth}
  \centering 
  \includegraphics[width=\textwidth]{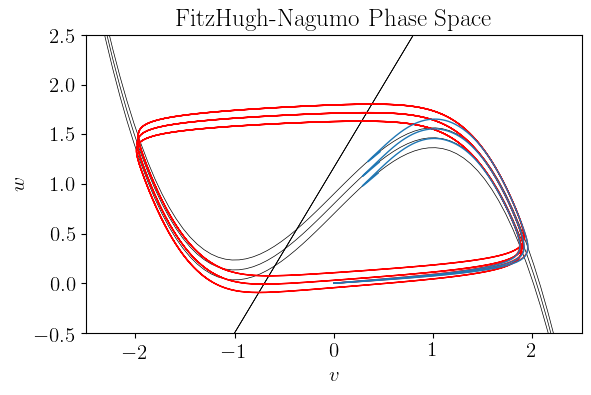}
  \caption{Without curriculum learning}
  \label{fig:fhnbadtrain}
    \end{subfigure}
  \begin{subfigure}{.5\textwidth}
    \centering
  \includegraphics[width=0.49\textwidth]{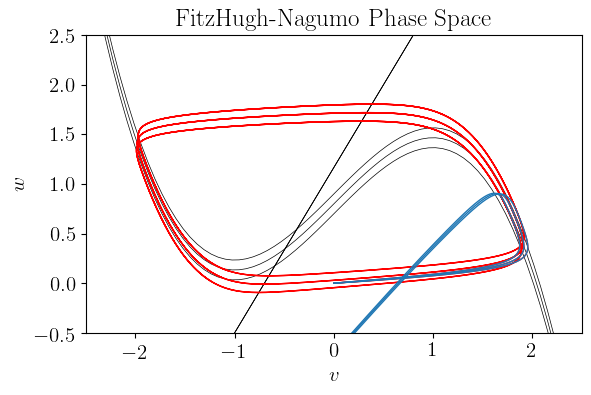}
  \includegraphics[width=0.49\textwidth]{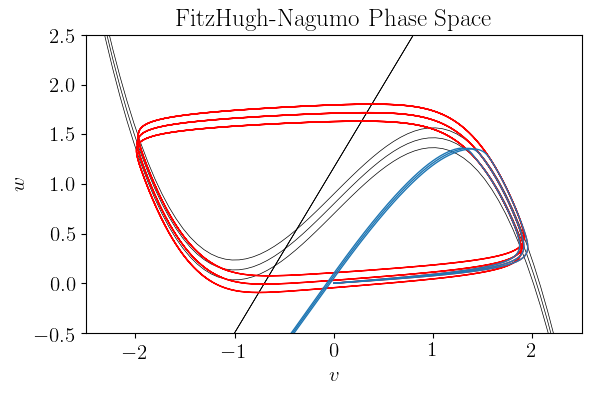}
  \includegraphics[width=0.49\textwidth]{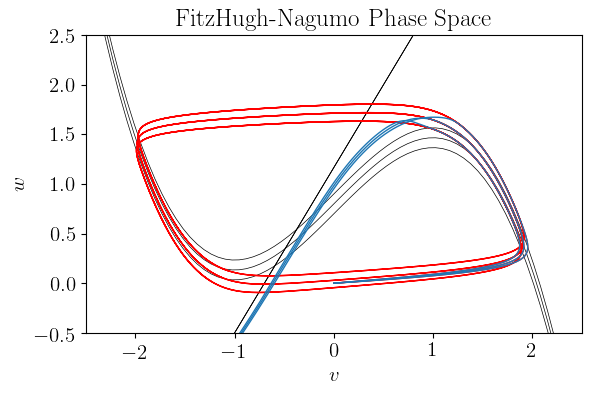}
  \includegraphics[width=0.49\textwidth]{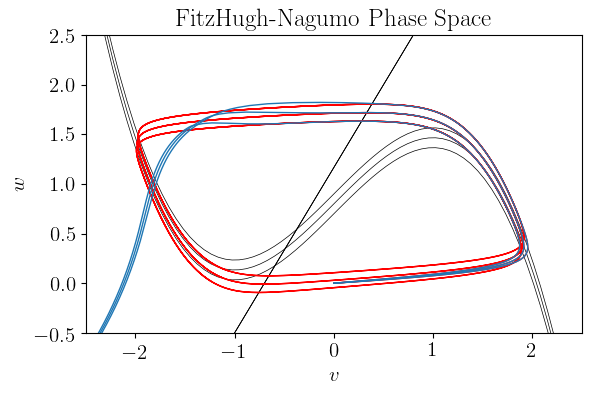}
  \caption{With curriculum learning}
  \label{fig:fhngoodtrain}
    \end{subfigure}
  \caption{Reference trajectories for three values of $I$ are shown in red, the corresponding nullclines for those values are shown in black, and the approximate solution bundle is shown in blue. Notice that without curriculum learning the neural network gets stuck on the cubic nullcline, while with curriculum learning this sticking point is safely passed during training.}
  \label{fig:fhnbothtrain}
\end{figure}

The FitzHugh-Nagumo model\cite{fitzhugh_impulses_1961,nagumo_active_1962} is a relaxation oscillator which can be used as a simple model of a biological neuron. Its state is described by a membrane voltage $v$, and a recovery variable $w$, and the ODE has parameters $a$, $b$, $\tau$, and $I$. The differential equation for the FitzHugh-Nagumo model is given by
\begin{align}
  \dv{v}{t} =  v - \frac{v^3}{3} - w + I,   \qquad
  \dv{w}{t} = \frac{1}{\tau} \qty(v + a - bw).\label{eq:fhneom1}
\end{align}
The nullclines of Eq. (\ref{eq:fhneom1}), \textit{i.e.} where $\dd{v}/\dd{t} = 0$ and $\dd{w}/\dd{t} = 0$, are plotted in black in Figure \ref{fig:fhnbothtrain}.


For this system, we found that it could be tricky to train the neural network solution bundle. When we simply sampled times uniformly from the time interval $[t_0, t_f]$, we found that the solution bundle could get ``stuck'' to the nullclines (as seen in Figure \ref{fig:fhnbadtrain}).

This seems to happen because while the approximate solution bundle is dragged around phase space during early training, the tail end can encounter a nullcline by chance, where it is easier to satisfy the ODE. Even though an error at an earlier time has to be made in the solution bundle for the tail to be on the nullcline, if the later times have very low error due to the ease of predicting a constant value whilst on the nullcline, the network weights can become stuck in this local minimum.

This pitfall can be avoided by applying curriculum learning\cite{bengio_curriculum_2009} to the training process. When training starts, we can restrict the time samples to come from $[t_0,t_m]$, where $t_m < t_f$ and $m$ is the batch number. As training progresses and the batch number $m$ is increased, we can make the task a bit more difficult by increasing $t_m$, requiring the network to learn a greater length of the solution bundle. The result of curriculum learning is shown in Figure \ref{fig:fhngoodtrain}. The entire solution bundle is plotted in each frame, but in the loss calculation only times up to $t_m$ are sampled. 

\section{Efficiency}

\begin{figure}[h!]
  \begin{subfigure}{.5\textwidth}
  \centering 
  \includegraphics[width=\textwidth]{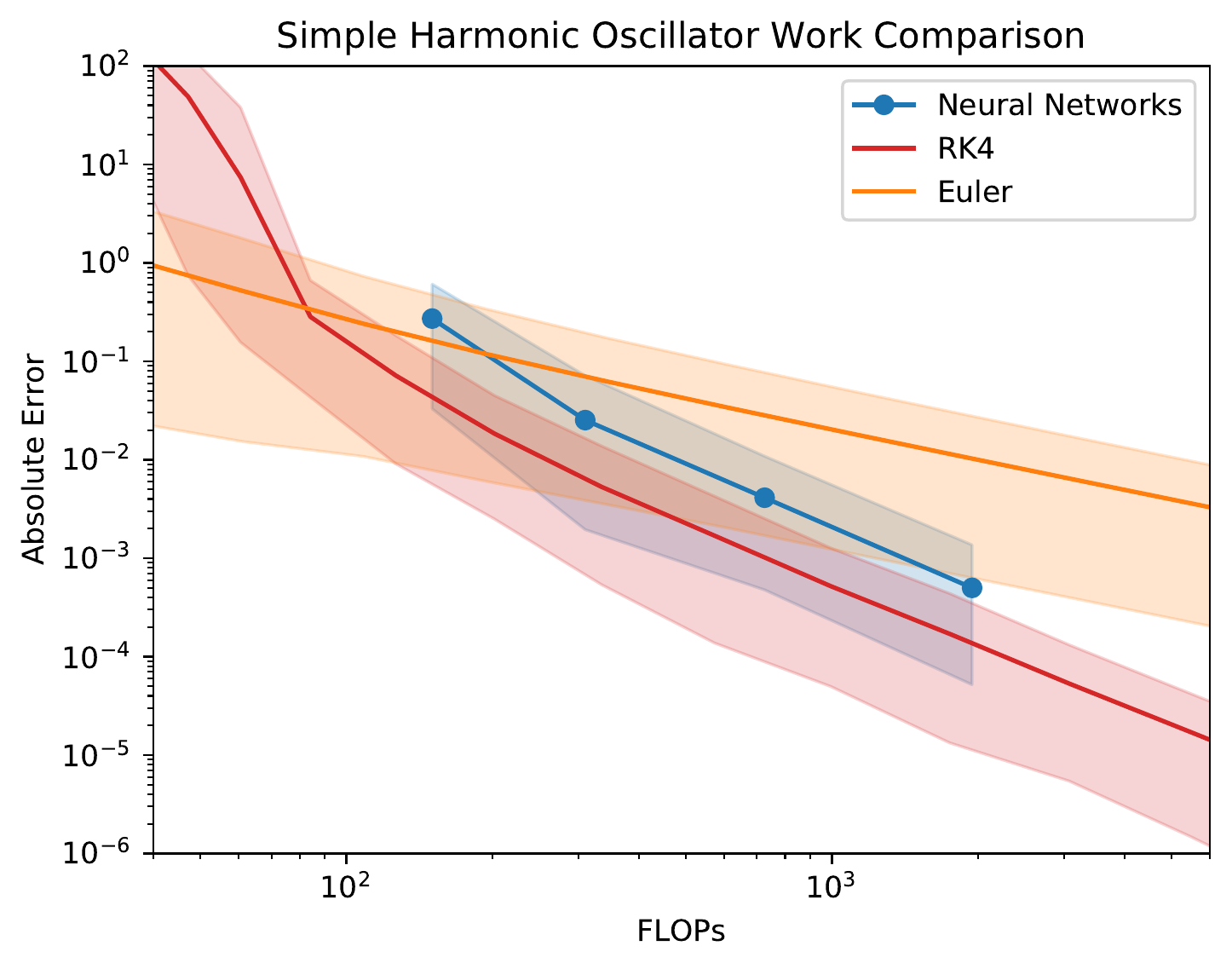}
  \caption{Computational Expense}
  \label{fig:shoefficienta}
\end{subfigure}
  \begin{subfigure}{.5\textwidth}
  \centering 
  \includegraphics[width=\textwidth]{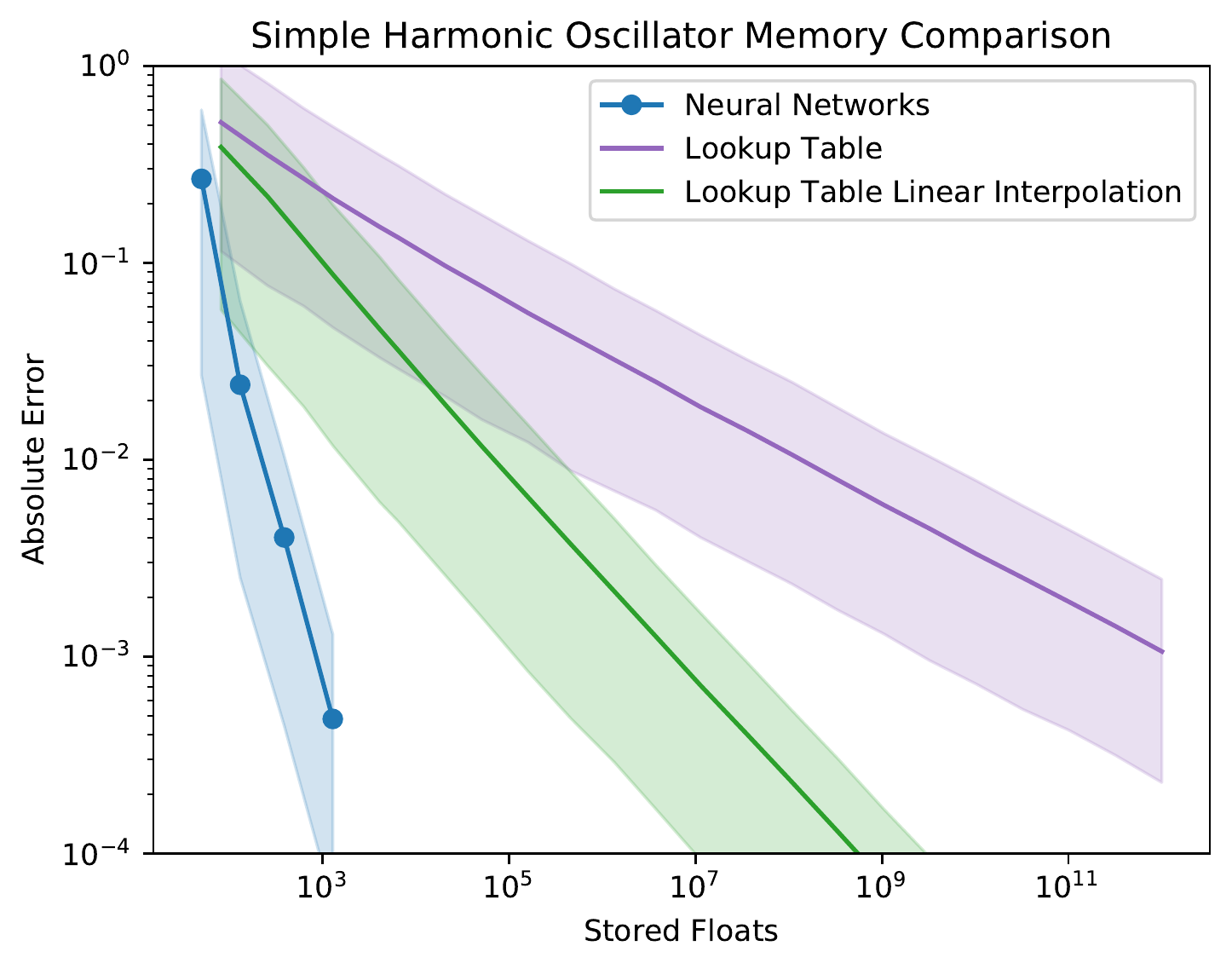}
  \caption{Memory Expense}
  \label{fig:shoefficientb}
\end{subfigure}
  \caption{Computational and memory efficiency comparisons of neural network solution bundles.}
\end{figure}

To get a sense of the computational and memory efficiency of neural network solution bundles, we compare them to Runge-Kutta and lookup tables, respectively. The equations of motion of a harmonic oscillator are given by
\begin{align}
  \dv{x}{t} = v, \qquad \dv{v}{t} = -\frac{k}{m} x,
  \label{eq:shoeom}
\end{align}
and we choose $m = 1$ for simplicity. We trained four dense neural networks with two hidden layers, with widths 4, 8, 16, and 32, over initial conditions $(x_0,v_0) \in [-1,1]\times[-1,1] $, parameters $k \in [0.5, 2]$, and times $[-0.01,2\pi]$. To evaluate the performance of these networks, we uniformly sampled points from $(x_0,v_0,k,t) \in [-1,1] \times [-1,1] \times [0.5,2] \times [0,2\pi]$ and computed the absolute error, which we define as $(\abs{\hat{x} - x} + \abs{\hat{v} - v})/2$, relative to the exact solution $(x,v)$ for each point. In Figure \ref{fig:shoefficienta}, the absolute errors for these same points are determined with Runge-Kutta (RK4) and the Euler method for increasingly small step sizes, and hence more floating point operations (FLOPs). The solid lines give the average absolute error and the bands encompass $90\%$ of the computed errors. While RK4 slightly outperforms the neural network approach, the operations in the evaluation of the dense neural network can be easily parallelized while RK4's steps cannot. As such, the simultaneous FLOPs per processor for the neural network can be kept nearly constant as the network complexity is increased, which can make the neural network solution bundle substantially faster in practice than a comparable RK4 approach, especially for an ODE with harder function evaluations than the multiplication in Eq. (\ref{eq:shoeom}).

Figure \ref{fig:shoefficientb} compares the neural network performances to simple uniformly-spaced lookup tables with the same number of divisions in $x_0$, $v_0$, $k$, and $t$. The neural network is a very compact representation of the solution bundle, achieving comparable performance to lookup tables taking up orders of magnitude more space. This is promising for the reusability of trained neural network solution bundles, as the smaller size makes it easier to share with others.

\section{Conclusion}

Backpropagation and the universal approximation theorem grants neural networks a unique ability to adapt and represent nearly any function. We extend the Lagaris method by introducing the concept of a neural network solution bundle, a group of approximate solutions to an ODE over a range of initial conditions and parameters. This allows for greater reuse of the trained network since it learns a variety of solutions. In addition, the solution bundle is differentiable in initial conditions and parameters, which can be useful for optimization tasks dependent on the value of the solution at given times. Other tasks that would require solving the differential equations repeatedly are also simplified, such as the propagation of uncertainty distributions across initial states, and for Bayesian inference in dynamical systems. The trained neural networks also have the potential to be used in low-power limited-memory settings that require quickly solving differential equations, as in drones or IoT devices. While the number of calculations involved in the training of neural network solution bundles is substantially higher than for computing a single solution using conventional methods, the cost can eventually be recouped if enough individual solutions are required, especially if the trained network is shared with other users. In addition, future advances in neural network training, evaluation, and specialized hardware will directly benefit this method. 


\section*{Broader Impact}

The work presented here has very general applicability since differential equations are widely employed in the sciences and engineering. As such, many ethical aspects of this research are neutral and primarily depend on specific applications in the future and their impacts. However, if optimally-trained networks for common differential equations are shared and made freely available, this method has the potential to reduce the cost of redundant calculations, helping with efficient energy use. If this work is used in systems that could affect lives or the environment, critical calculations should be checked against other numerical approaches for safety.

\begin{ack}

  We would like to thank Marios Mattheakis and Michael Emanuel for helpful discussions which refined the ideas in this work. The computations in this paper were run on the FASRC Cannon cluster supported by the FAS Division of Science Research Computing Group at Harvard University.
\end{ack}

\normalsize
\bibliographystyle{unsrt}
\bibliography{nnbundle}

\clearpage

\section{Appendix}
\begin{appendices}

  \section{Proof of Global Error Bound}

One appeal for using a decaying exponential weighting comes from comparison to one-step methods like Runge-Kutta. Given a differential equation of the form
\begin{align}
  \dv{x}{t} = f(t,x),
  \label{eq:simplediffeq}
\end{align}
a one-step method will generate a sequence through 
\begin{align}
  x_{n+1} = x_n + h\Phi(t_n, x_n;h),
\end{align}
where the second term is the increment, which for example in the Euler method would be $h\Phi(t_n, x_n ;h) = h f(t_n, x_n)$. The truncation (local) error is given by 
\begin{align}
  \varepsilon_n = \frac{x(t_{n+1}) - x(t_n)}{h} - \Phi(t_n, x(t_n); h), \qqquad h = t_{n+1} - t_n.
  \label{eq:truncationerror}
\end{align}
The global error in a one-step method, $\epsilon_N = x(t_N) - x_N$, which is due to the accumulated truncation error, is bounded by\cite{suli_introduction_2003}
\begin{align}
  \abs{\epsilon_N} \leq \frac{\varepsilon}{L_{\Phi}}\qty(e^{L_{\Phi}(t_N - t_0)} - 1)
  \label{eq:globaltruncerror}
\end{align}
where $\varepsilon = \max_{0\leq n \leq N-1} \abs{\varepsilon_n}$ and, $\Phi$ is Lipschitz continuous with Lipschitz constant $L_\phi$, $\abs{\Phi(t,u;h) - \Phi(t,v;h) } \leq L_{\Phi}\abs{u-v}$. 

We can derive an analogue of this bound on global error for the neural network solution bundle. The local error of Eq. (\ref{eq:truncationerror}) should be compared to
\begin{align}
  \bm{\varepsilon}(t_i \: ; \: \vb{x}_{0i}, \bm{\theta}_i) \equiv \vb{G}\qty(\hat{\vb{x}}(t_i \: ; \: \vb{x}_{0i}, \bm{\theta}_i), \pdv{\hat{\vb{x}}(t_i \: ; \: \vb{x}_{0i}, \bm{\theta}_i)}{t}, t_i \: ; \: \bm{\theta}_i),
  \label{eq:appendixlocalerror}
\end{align}
which for a differential equation of the form in Eq. (\ref{eq:simplediffeq}) would give
\begin{align}
  \varepsilon(t) = \dv{\hat{x}(t)}{t} - f(t,\hat{x}).
  \label{eq:appendixlocalerrorsimp}
\end{align}
In our approach, Eq. (\ref{eq:appendixlocalerrorsimp}) is effectively a truncation error for an infinitesimal timestep. The global error $\epsilon(t)$ is given by
\begin{align}
  \epsilon(t) = \hat{x}(t) - x(t),
\end{align}
which we can substitute into Eq. (\ref{eq:appendixlocalerrorsimp}) to obtain
\begin{align}
  \dv{\epsilon}{t} = \varepsilon(t) - \dv{x}{t} + f\qty\big(t, x(t) + \epsilon(t)).
\end{align}
Assuming $f$ is a Lipschitz-continuous function, $\abs\Big{f\qty\big(t, x(t) + \epsilon(t)) - f\qty\big(t, x(t))} \leq L_f \abs{\epsilon(t)}$, so the absolute value of the right-hand side  can be written
\begin{align}
  \abs{\varepsilon(t) - \dv{x}{t} + f\qty\big(t, x(t) + \epsilon(t))} &=  \abs\Big{\varepsilon(t) + f\qty\big(t, x(t) + \epsilon(t)) - f\big(t, x(t))}\\
  &\leq \abs{\varepsilon(t)} + L_f \abs{\epsilon(t)} \\
  &\leq \varepsilon_{t'} + L_f\abs{\epsilon(t)},
\end{align}
where we have made use of the differential equation $\dd{x}/\dd{t} = f\big(t,x(t))$, the triangle inequality, and $\varepsilon_{t'} = \max_{t_0 \leq t \leq t'} \abs{\varepsilon(t)}$. Since $\epsilon(t_0) = 0$, to find a bound on $\abs{\epsilon(t)}$ we can consider the solution to the ODE for upper bound $E(t) \geq \abs{\epsilon(t)} \geq 0$,
\begin{align}
  \dv{E}{t} &= \varepsilon_{t'} + L_f E(t)
\end{align}
where $E(t_0) = 0$, which is $E(t) = \frac{\varepsilon_{t'}}{L_f} \qty(\exp \qty[L_f \qty(t - t_0)] - 1)$. Thus,
\begin{align}
  \abs{\epsilon(t)} \leq \frac{\varepsilon_{t'}}{L_f} \qty(e^{L_f \qty(t - t_0)} - 1),
  \label{eq:appendixglobalerror}
\end{align}
which can be compared to the global error bound of the discrete case, Eq. (\ref{eq:globaltruncerror}).

\section{Propagating a Probability Distribution}

A neural network solution bundle provides a mapping from initial conditions to the state at later times. This can be useful for time-evolving a distribution over initial conditions to obtain a probability distribution over states at later time $t$. Given a probability density over initial states $p_{0}(\vb{x}_0)$, we note that the solution bundle $\vb{x}(t \: ; \: \vb{x}_0)$ at time $t$ describes a coordinate transformation from $\vb{x}_0$, transforming the coordinates $\vb{x}_0$ to $\vb{x}_t$. If this transformation is invertible and differentiable from the subset of initial state space $X_0$ to the final state space $X_t$, we can write out the probability density of later states, 
\begin{align}
  p_{t}\qty(\vb{x}_t) = p_{0}\qty(\vb{f}^{-1}(\vb{x}_t)) \abs{\vb{J}_{\vb{f}^{-1}}} ,
  \label{eq:probabilitytransform}
\end{align}
where $\vb{f}^{-1}(\vb{x}_t)$ is the inverse of $\vb{f}(\vb{x}_0) \equiv \vb{x}(t \: ; \: \vb{x}_0)$, and $\vb{J}_{\vb{f}^{-1}} = \pdv{\vb{x}_0}{\vb{x}_t}$ is the Jacobian of $\vb{f}^{-1}$.
The opposite task, \textit{i.e} where a probability distribution over later states $p_t(\vb{x}_t)$ is known and the distribution over initial states is desired, takes on an even more convenient form where the neural network solution bundle does not have to be inverted:
\begin{align}
  p_{0}\qty(\vb{x}_0) = p_{t}\qty(\vb{f}(\vb{x}_0)) \abs{\vb{J}_{\vb{f}}},
  \label{eq:probabilitytransformreverse}
\end{align}
where $\vb{J}_{\vb{f}} = \pdv{\vb{x}_t}{\vb{x}_0}$ is the Jacobian of $\vb{f}$, which can be calculated exactly using automatic differentiation of the solution bundle. This gives a closed analytic form for the desired probability density. Note that if the ODE is time-reversible, Eq. (\ref{eq:probabilitytransform}) can be converted to the easier form Eq. (\ref{eq:probabilitytransformreverse}) by simply treating the target space as the input space and training the solution bundle on the time-reversed equations of motion.

In the case of an energy-conserving Hamiltonian dynamical system where all the state variables are canonical coordinates, Liouville's theorem guarantees that $\abs{\vb{J}_{\vb{f}}} = \abs{\vb{J}_{\vb{f}^{-1}}}$, resulting in particularly simple relationships for the probability densities at different times.

\subsection{Planar Circular Restricted Three-Body Problem}

If we have two measurements of the position of an asteroid at two different times, along with some uncertainty,
\begin{subequations}
\begin{align}
  (t_0, x_0, y_0) = (0.00, 1.0510 \pm 0.0003, 0.1000 \pm 0.0003) \\
  (t_1, x_1, y_1) = (0.05, 1.0276 \pm 0.0003, 0.0878 \pm 0.0003)
\end{align}
\end{subequations}
we can compute the probability distribution of future positions with the neural network solution bundle. Let $p\qty\big(\vb{r}(t) = (x,y)^\T \mid \qty{\vb{r}_1, \vb{r}_0})$ be the probability density of the position being $(x,y)^\T$ at time $t$, given the position measurements $\vb{r}_0$ and $\vb{r}_1$. By marginalizing over the final velocities, we obtain
\begin{align}
  p\qty\big(\vb{r}(t) = (x,y)^\T \mid \qty{\vb{r}_1, \vb{r}_0}) &= \iint p\qty\big(\vb{q}(t) = (x,y,u,v)^\T \mid \qty{\vb{r}_1, \vb{r}_0}) \dd{u} \dd{v}.
\end{align}
To compute the integrand, we can use Bayes' theorem,
\begin{align}
  p\qty\big(\vb{q}(t) =  (x,y,u,v)^\T \mid \qty{\vb{r}_1, \vb{r}_0}) &=  \frac{p\qty\big(\qty{\vb{r}_1, \vb{r}_0} \mid \vb{q}(t) = (x,y,u,v)^\T) p\qty\big(\vb{q}(t) = (x,y,u,v)^\T)}{p\qty\big(\qty{\vb{r}_1, \vb{r}_0})} \nonumber \\
   &\propto  p\qty\big(\vb{r}_1 \mid \vb{q}(t) = (x,y,u,v)^\T)p\qty\big(\vb{r}_0 \mid \vb{q}(t) = (x,y,u,v)^\T),
   \label{eq:earthmoonsampleweight}
\end{align}
where in the last step we have assumed a uniform prior, and that the errors in the two position measurements $\vb{r}_0$ and $\vb{r}_1$ are independent.

\begin{figure}[h]
  \centering 
  \includegraphics[width=0.49\textwidth]{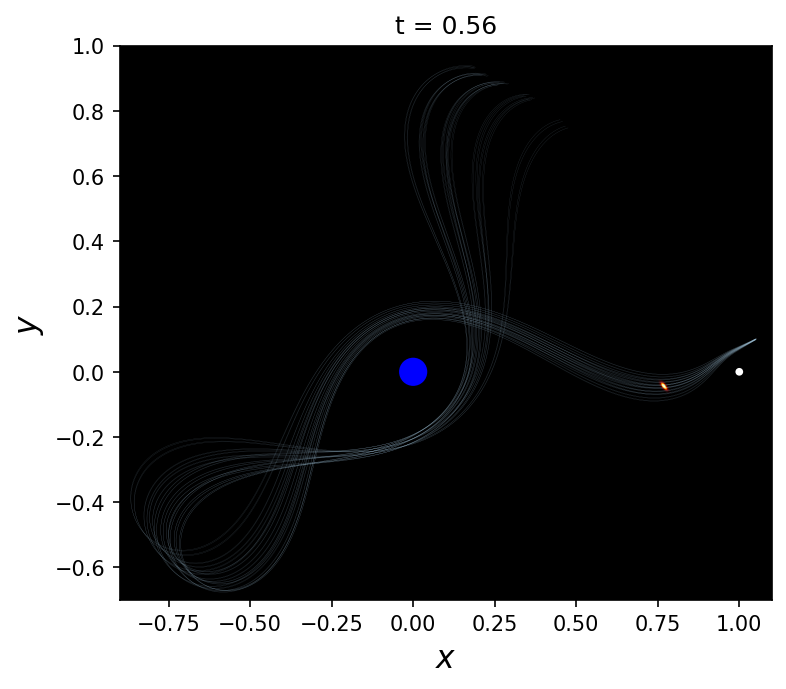}
  \includegraphics[width=0.49\textwidth]{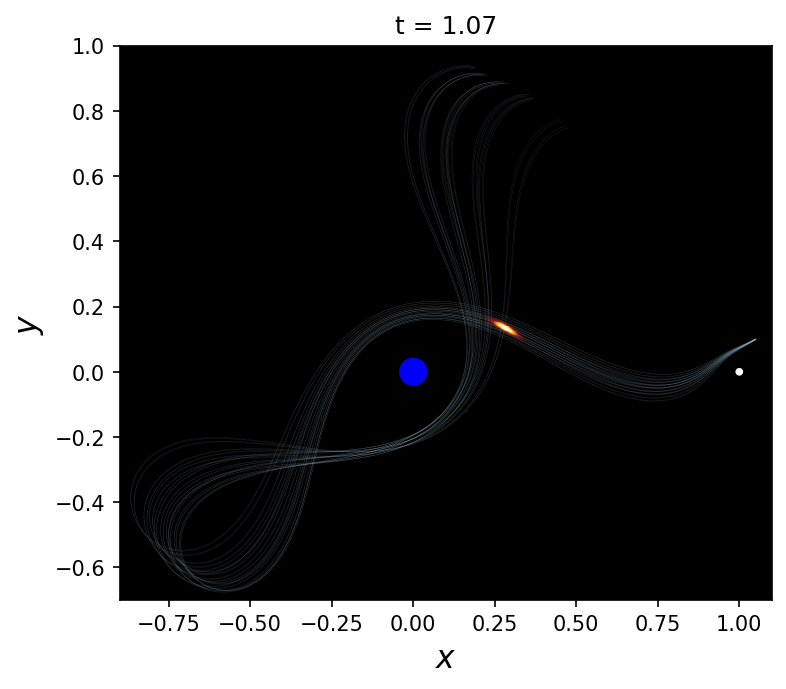}
  \includegraphics[width=0.49\textwidth]{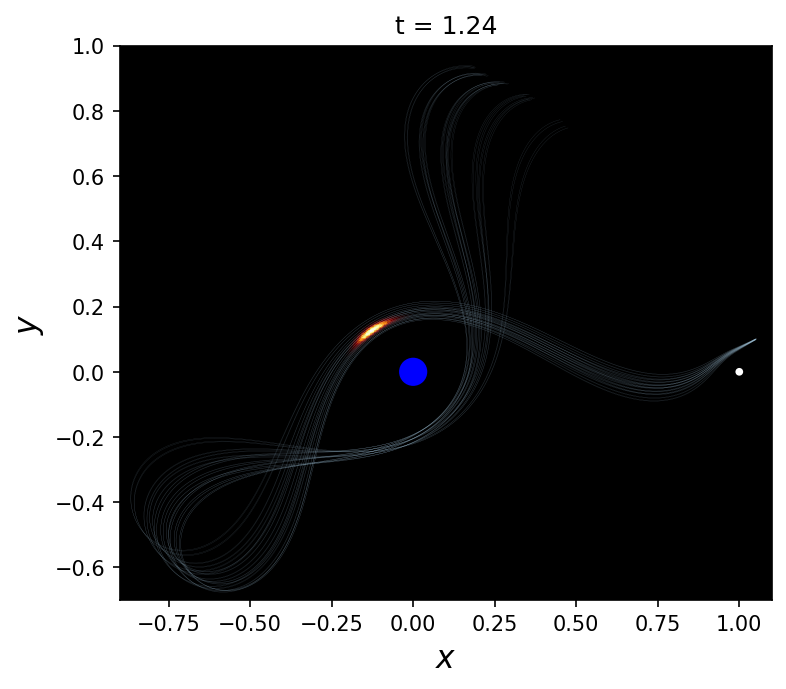}
  \includegraphics[width=0.49\textwidth]{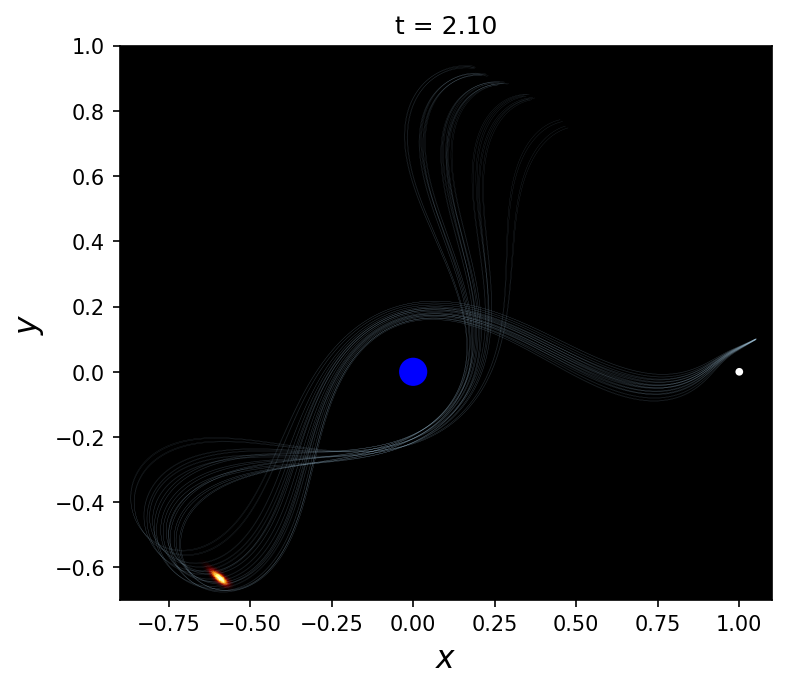}
  \includegraphics[width=0.49\textwidth]{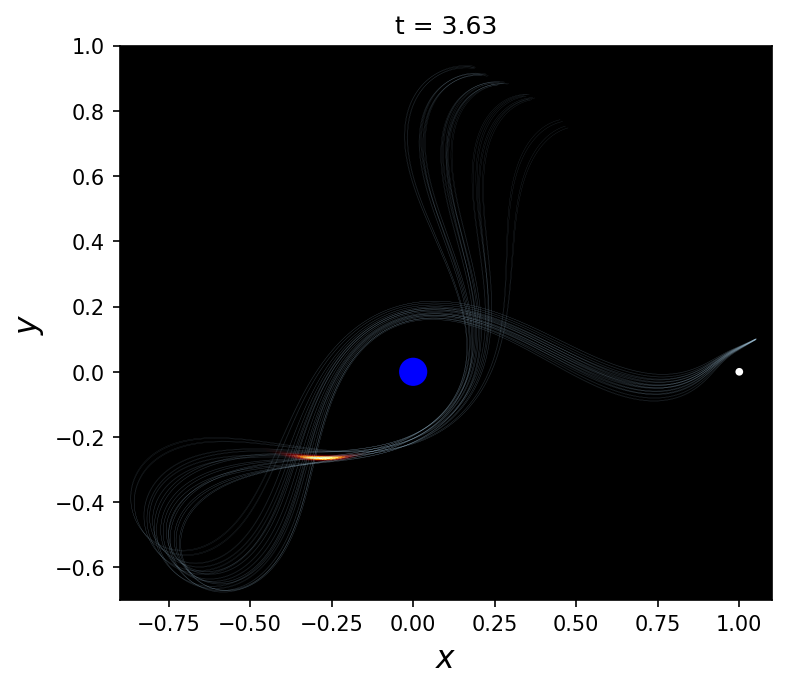}
  \includegraphics[width=0.49\textwidth]{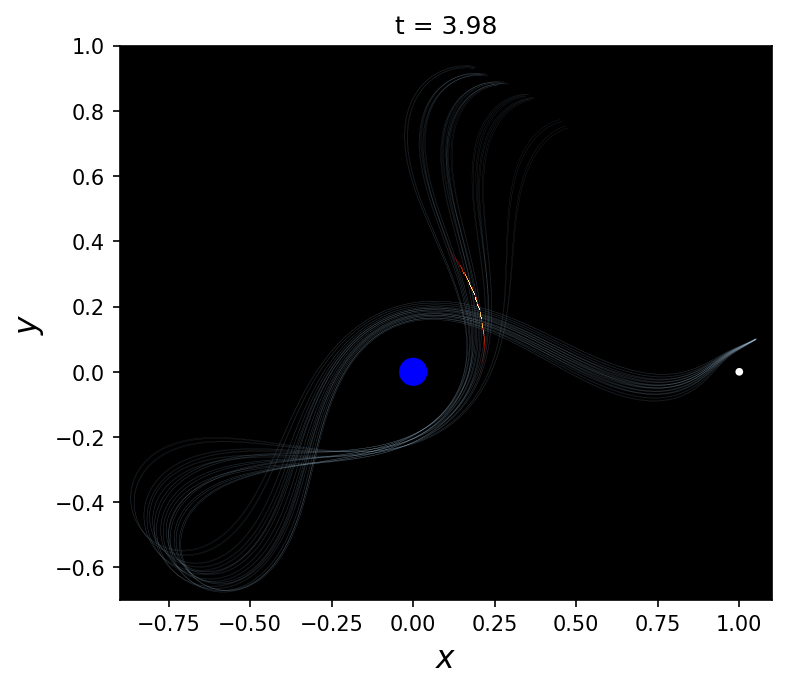}
  \caption{Probability distribution $p\qty(\vb{x}(t) = (x,y)^\T)$ at various times. A few trajectories in the bundle are shown in white, and the distribution is shown as a heatmap.}
  \label{fig:earthmoondistappendix}
\end{figure}
Iterating over a uniform grid of initial positions and velocities, evaluating the solution at time $t_1$ and $t$, weighting the samples with the probability densities given by Eq. (\ref{eq:earthmoonsampleweight}), and forming a weighted histogram of positions, we have the approximate distribution of the asteroid's location at time $t$. Figure \ref{fig:earthmoondistappendix} shows the position distribution at various final times $t$. When uniform sampling becomes infeasible due to high dimensionality of the inputs, it is straightforward to use MCMC instead.

\section{Computational Details}

All calculations were performed with PyTorch 1.3.1 on a single node equipped with a GPU:

\begin{itemize}
  \item Model name: Intel Xeon Gold 6126 CPU @ 2.60GHz
  \item CPUs: 24
  \item RAM: 191898 MB
  \item GPU: Nvidia Tesla V100-PCIE-16GB
\end{itemize}

\subsection{Planar Circular Restricted Three-Body Problem}

\textbf{ODE:}
\begin{align}
  &\dv{x}{t} = u\;, \qquad \dv{u}{t} = x - \mu + 2v - \qty[ \frac{\mu(x-1)}{\qty\big( \qty(x-1)^2 + y^2)^{3/2}} + \frac{(1 - \mu) x}{\qty(x^2 + y^2)^{3/2}}],& \\
  &\dv{y}{t} = v\;, \qquad \dv{v}{t} = y - 2u - \qty[  \frac{\mu y}{\qty\big( \qty(x-1)^2 + y^2)^{3/2}} + \frac{\qty(1 - \mu) y}{ \qty(x^2 +y^2)^{3/2}}].&
  \label{eq:appendixcrtbEOM}
\end{align}
The above equation is in the co-rotating frame in the plane of the two heavy bodies (Earth and Moon) about their barycenter. The position of the third body of negligible mass is given by $(x,y)$, with velocity $(u,v)$. The parameter $\mu$ is the ratio of the Moon mass to the Earth plus Moon combined mass.

\begin{table}[h]
  \begin{tabular}{lll}
    \toprule
    \multicolumn{3}{l}{Network Architecture} \\
    \midrule
  Input layer & 5 neurons & $\tanh$\\
  Hidden layers & 8 dense layers of 128 neurons each & $\tanh$\\
  Output layer & 4 neurons & linear\\
 \bottomrule
 \end{tabular}
 \caption{Planar circular restricted three-body problem network architecture.}
\end{table}

\begin{table}[h]
  \begin{tabular}{lll}
    \toprule
    Quantity & Value \\
    \midrule
 Weighting function & $b(t) = \exp(-2 t)$  & \\
 Initial condition ranges & \multicolumn{2}{l}{$(x_0,y_0,u_0,v_0) \in [1.05,1.052]\times[0.099, 0.101] \times [-0.5, -0.4] \times [-0.3, -0.2]$}  \\
ODE parameter & $\mu = 0.01$ & \\
Time range & $[-0.01, 5]$ & \\
Optimizer & Adam & \\
Batch size & 10,000 & \\
Learning rate & batch 0 to 3,999,999 & $\eta=0.001$, ReduceLROnPlateau \\
& &  \quad factor=0.5, patience=200000, threshold = 0.5, \\
& &  \quad threshold\_mode = 'rel', cooldown = 0, \\
& &  \quad  min\_lr = 1e-6, eps = 1e-8 \\
& batch 4,000,000 to 7,999,999 & $\eta=0.000001$ \\
& batch 8,000,000 to 35,999,999 & $\eta=0.00001$ \\
Training rate & 56 batches/sec & \\
Training time & 178.6 hours & \\
 \bottomrule
 \end{tabular}
 \caption{Planar circular restricted three-body problem training hyperparameters and other details.}
\end{table}

\begin{figure}[h]
  \begin{subfigure}{.5\textwidth}
  \centering 
  \includegraphics[width=\textwidth]{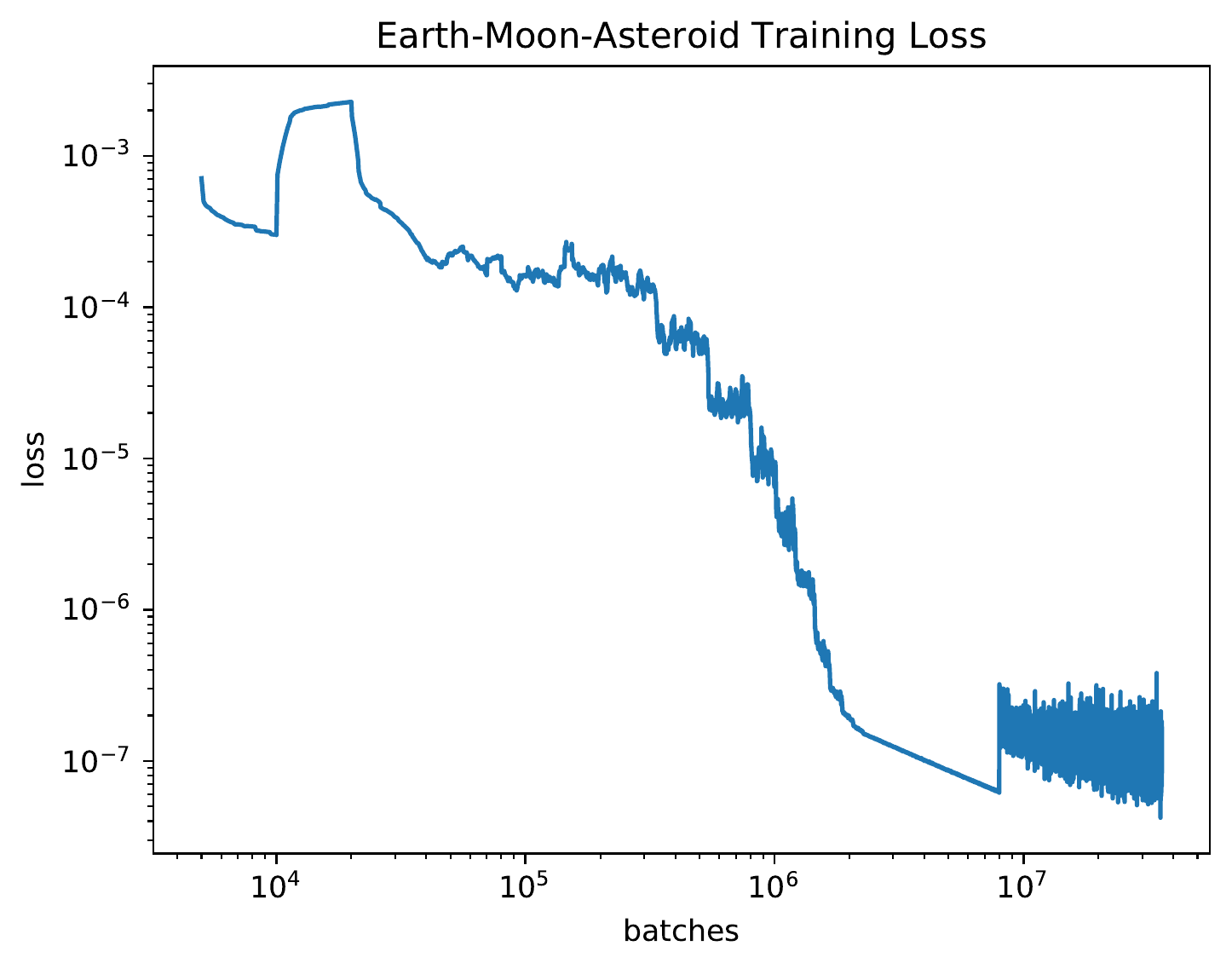}
\end{subfigure}
  \begin{subfigure}{.5\textwidth}
  \centering 
  \includegraphics[width=0.49\textwidth]{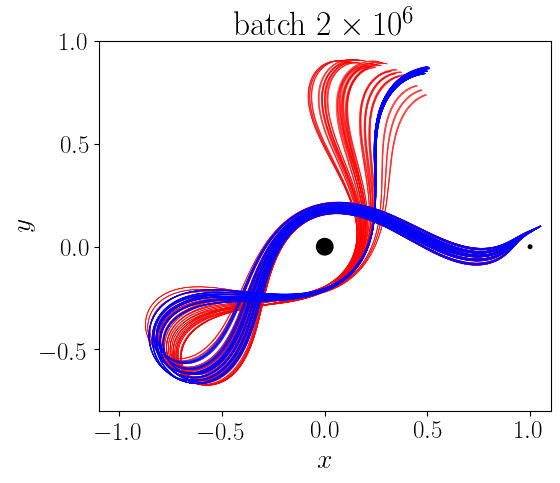}
  \includegraphics[width=0.49\textwidth]{figures/crtb/training/00398.png}
  \includegraphics[width=0.49\textwidth]{figures/crtb/training/00500.png}
  \includegraphics[width=0.49\textwidth]{figures/crtb/training/00727.png}
\end{subfigure}
\caption{\textbf{Left:} Loss during training for the planar circular restricted three-body problem, smoothed with a moving average of window size 10,000. \textbf{Right:} Solution bundle after select batches during training. These are presented to demonstrate why we increased the learning rate starting at batch $8 \times 10^6$, causing the temporary increase in the loss. At batch $8 \times 10^6$ notice that the tail of the solution bundle is not fanning out, and it appears that the neural network is trapped in a local minimum. Increasing the learning rate helped the network get out, and the solution bundle once again began approaching the reference trajectories in red.}
  \label{fig:crtbloss}
\end{figure}

We found that reducing the learning rate during training greatly helped in decreasing the loss for this system. In Figure \ref{fig:crtbloss} every time the learning rate was halved, the loss immediately decreased, forming downward steps. However, the learning rate was made too small through this process, and the network approached a suboptimal minimum by batch $8 \times 10^6$. Curiously, after increasing the learning rate, despite temporarily increasing the loss, the solution bundle immediately started converging better to the true trajectories determined with Runge-Kutta. This is due to the loss assessing the solution bundle's satisfaction of the ODE instead of its accuracy compared to the true solution. As seen in Figure \ref{fig:crtbloss}, if small errors are made at early times in the solution bundle such that the neural network has an easier task at later times, the earlier errors might not be corrected. This is seen by the trajectories at later times not fanning out.

\subsection{Rebound Pendulum}

\textbf{ODE:}
\begin{align}
  \dv{\theta}{t} = \omega, \qquad \dv{\omega}{t} = - \frac{g}{\ell} \sin \theta + H(-\theta) \trm{ReLU}\qty(-\frac{k\ell}{m} \theta  - c \omega),
  \label{eq:reboundpendulumeom}
\end{align}
where $H(x)$ is the Heaviside step function, and $\trm{ReLU}(x)$ is the rectifier function. The position of the pendulum is described by its angle $\theta$, and its angular velocity is $\omega$. The parameters in the equation are: gravitational acceleration $g$, pendulum length $\ell$, spring constant $k$, damping coefficient $c$, and mass $m$. For simplicity we set $g$, $\ell$, and $m$ to unity. The second term in the second differential equation describes the interaction of the spring with the pendulum, capturing the fact that the spring only interacts with the pendulum for negative angles and that it can only push the pendulum away.

\begin{table}[h]
  \begin{tabular}{lll}
    \toprule
    \multicolumn{3}{l}{Network Architecture} \\
    \midrule
  Input layer & 5 neurons & $\tanh$ \\
  Hidden layers & 8 dense layers of 128 neurons each & $\tanh$ \\
  Output layer & 2 neurons & linear\\
 \bottomrule
 \end{tabular}
 \caption{Rebound pendulum network architecture.}
\end{table}

\begin{table}[h]
  \begin{tabular}{lll}
    \toprule
    Quantity & Value \\
    \midrule
 Weighting function & $b(t) = \exp(-0.5 t)$  & \\
 Initial condition ranges & \multicolumn{2}{l}{$(\theta_0,\omega_0) \in [0.0,1.0]\times[-0.2, 0.2]$}  \\
 ODE parameter ranges & \multicolumn{2}{l}{$(k,c) \in [2.0,5.0]\times[0.0, 2.0]$}  \\
Time range & $[-0.01, 10]$ & \\
Optimizer & Adam & \\
Batch size & 10,000 & \\
Learning rate & batch 0 to 1,999,999 & $\eta=0.001$, ReduceLROnPlateau \\
& &  \quad factor=0.5, patience=1500000, threshold = 0.5, \\
& &  \quad threshold\_mode = 'rel', cooldown = 0, \\
& &  \quad  min\_lr = 0, eps = 1e-8 \\
& batch 0 to 1,999,999 & $\eta=0.00001$, ReduceLROnPlateau \\
& &  \quad factor=0.5, patience=2000000, threshold = 0.5, \\
& &  \quad threshold\_mode = 'rel', cooldown = 0, \\
& &  \quad  min\_lr = 0, eps = 1e-8 \\
Training rate & 98 batches/sec & \\
Training time & 11.37 hours & \\
 \bottomrule
 \end{tabular}
 \caption{Rebound pendulum training hyperparameters and other details.}
\end{table}

\begin{figure}[h]
  \centering 
  \includegraphics[width=0.5\textwidth]{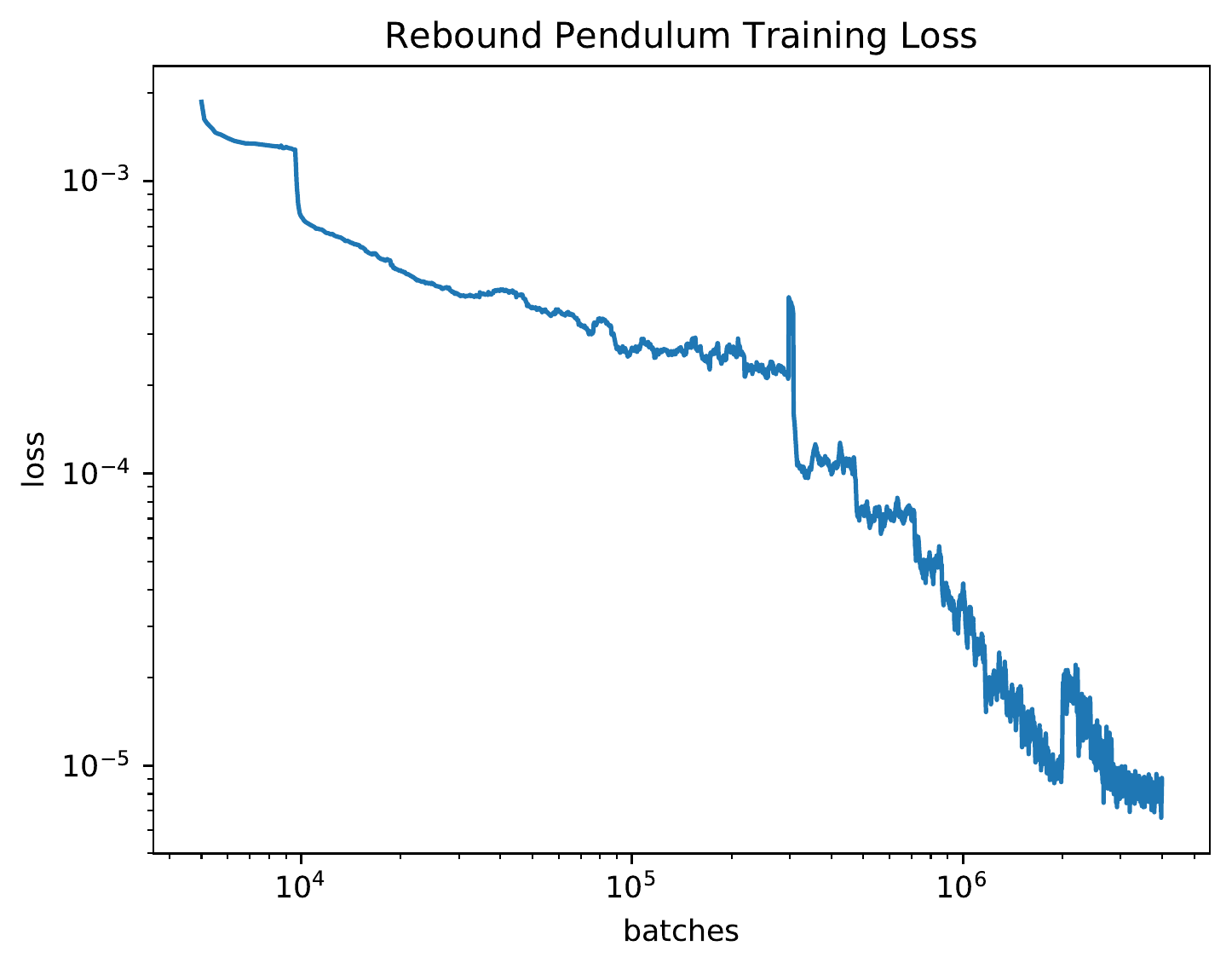}
\caption{Rebound pendulum loss during training, smoothed with a moving average of window size 10,000.}
  \label{fig:reboundpendulumloss}
\end{figure}

\clearpage

\subsection{FitzHugh-Nagumo Model}

\textbf{ODE:}
\begin{align}
  \dv{v}{t} =  v - \frac{v^3}{3} - w + I,   \qquad
  \dv{w}{t} = \frac{1}{\tau} \qty(v + a - bw).
\end{align}
The two state variables are $v$ and $w$ which are membrane voltage and recovery, respectively. The parameters of the system are $a$, $b$, which control the shape of the trajectory, $\tau$, which sets the timescale, and $I$, the external stimulus. The nullclines of the ODE are
\begin{align}
  w &= v - \frac{v^3}{3} + I, &\Rightarrow \dv{v}{t} = 0 \\
  w &= \frac{v+a}{b}, &\Rightarrow \dv{w}{t} = 0
\end{align}
which are helpful for understanding the direction reversals in each state variable of the trajectory in phase space. 

\begin{table}[h]
  \begin{tabular}{lll}
    \toprule
    \multicolumn{3}{l}{Network Architecture} \\
    \midrule
  Input layer & 7 neurons & $\tanh$ \\
  Hidden layers & 4 dense layers of 128 neurons each & $\tanh$ \\
  Output layer & 2 neurons & linear\\
 \bottomrule
 \end{tabular}
 \caption{FitzHugh-Nagumo (without curriculum learning) network architecture.}
\end{table}

\begin{table}[h]
  \begin{tabular}{lll}
    \toprule
    Quantity & Value \\
    \midrule
 Weighting function & $b(t) = \exp\qty[- \lambda_m t]$  & \\
 & $\lambda_m = \exp\qty[ - \ln\qty(100) \frac{m}{M} ]$ & where $m$ is the batch number, $M$ is total batches \\
 Initial condition ranges & \multicolumn{2}{l}{$(v,w) \in [-0.2,0.2]\times[-0.2, 0.2]$}  \\
 ODE parameter ranges & \multicolumn{2}{l}{$(a,b,\tau,I) \in [0.6,0.9]\times[0.6, 0.9] \times [10., 14.] \times [0.6, 1.0]$}  \\
Time range & $[-0.10, 100]$ & \\
Optimizer & Adam & \\
Batch size & 10,000 & \\
Learning rate & batch 0 to 499,999 & $\eta=0.001$ \\
Training rate & 151 batches/sec & \\
Training time & 55 minutes & \\
 \bottomrule
 \end{tabular}
 \caption{FitzHugh-Nagumo (without curriculum learning) training hyperparameters and other details.}
\end{table}

\begin{figure}[h]
  \centering 
  \includegraphics[width=0.49\textwidth]{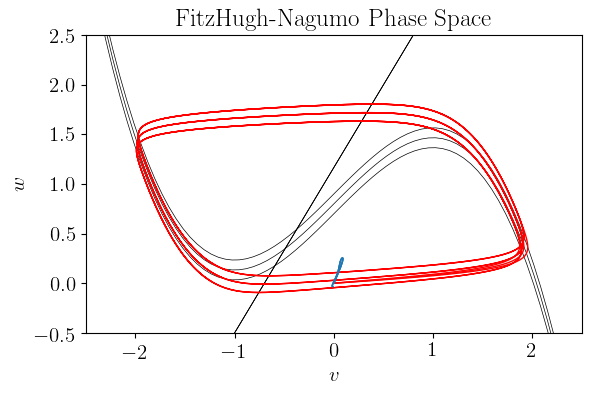}
  \includegraphics[width=0.49\textwidth]{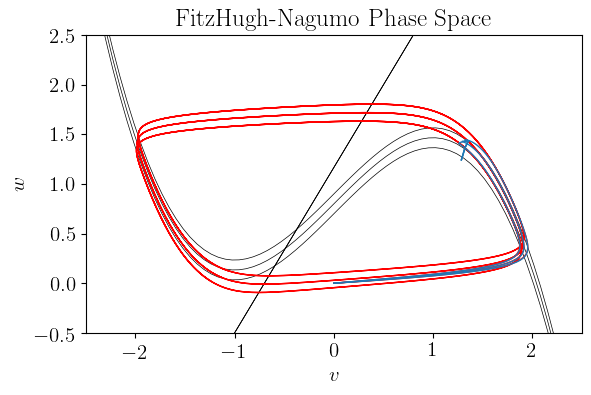}
  \includegraphics[width=0.49\textwidth]{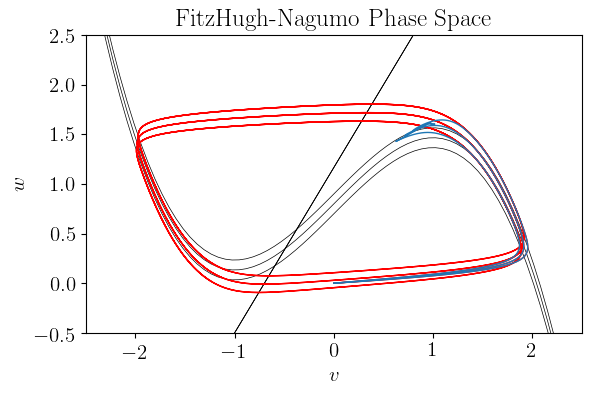}
  \includegraphics[width=0.49\textwidth]{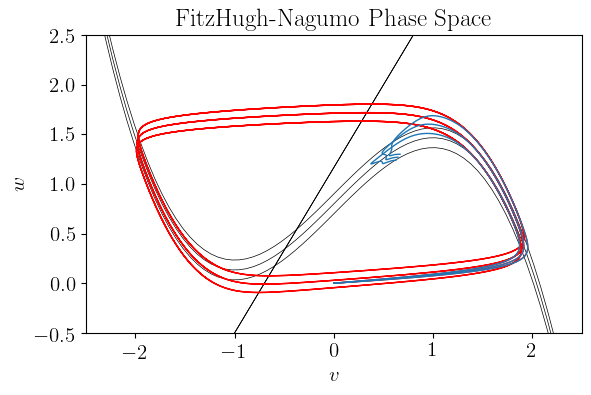}
  \includegraphics[width=0.49\textwidth]{figures/fhn/badtrain/00040.png}
  \includegraphics[width=0.49\textwidth]{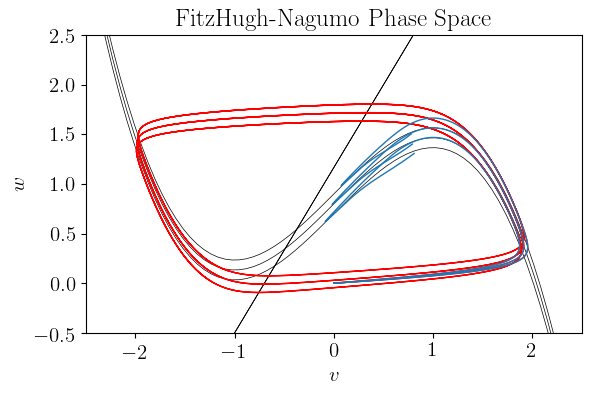}
  \caption{Training FitzHugh-Nagumo solution bundles without curriculum learning. The trajectories bind to the nullclines early on and the solution bundle does not converge properly.}
  \label{fig:fhnbadappendix}
\end{figure}

\begin{table}[h]
  \begin{tabular}{lll}
    \toprule
    \multicolumn{3}{l}{Network Architecture} \\
    \midrule
  Input layer & 7 neurons & $\tanh$ \\
  Hidden layers & 8 dense layers of 121 neurons each & $\tanh$ \\
   & skip connections from input concatenated to output of every hidden layer & linear \\
  Output layer & 2 neurons & linear\\
 \bottomrule
 \end{tabular}
 \caption{FitzHugh-Nagumo (curriculum learning) network architecture.}
\end{table}

\begin{table}[h]
  \begin{tabular}{lll}
    \toprule
    Quantity & Value \\
    \midrule
 Dynamic end time & $t_m = \frac{100}{\ln(11)} \ln\qty(10\frac{m}{M} + 1) $  & where $m$ is the batch number, $M$ is total batches\\
 Weighting function & $b(t) = \exp\qty[- \lambda_m t]$  & \\
 & $\lambda_m = \frac{4}{t_m + 5}$ &  \\
 Initial condition ranges & \multicolumn{2}{l}{$(v,w) \in [-0.1,0.1]\times[-0.1, 0.1]$}  \\
 ODE parameter ranges & \multicolumn{2}{l}{$(a,b,\tau,I) \in [0.6,0.8]\times[0.5, 0.7] \times [11., 14.] \times [0.7, 0.9]$}  \\
Time range & $[-0.10, 100]$ & \\
Optimizer & Adam & \\
Batch size & 10,000 & \\
Learning rate & batch 0 to 9,999,999 & $\eta=0.0001$ \\
Training rate & 80 batches/sec & \\
Training time & 34.8 hours & \\
 \bottomrule
 \end{tabular}
 \caption{FitzHugh-Nagumo (curriculum learning) training hyperparameters and other details.}
\end{table}

\begin{figure}[h]
  \centering 
  \includegraphics[width=0.45\textwidth]{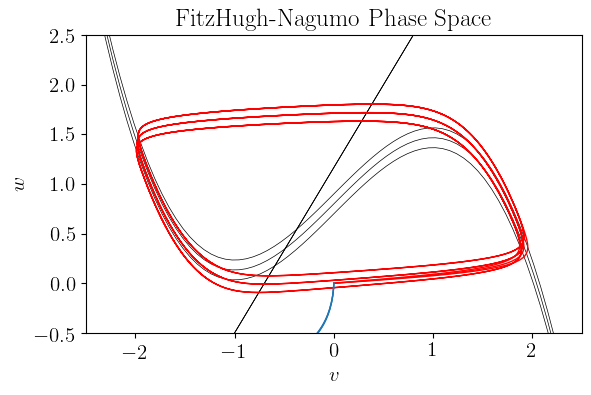}
  \includegraphics[width=0.45\textwidth]{figures/fhn/goodtrain/00001.png}
  \includegraphics[width=0.45\textwidth]{figures/fhn/goodtrain/00002.png}
  \includegraphics[width=0.45\textwidth]{figures/fhn/goodtrain/00003.png}
  \includegraphics[width=0.45\textwidth]{figures/fhn/goodtrain/00004.png}
  \includegraphics[width=0.45\textwidth]{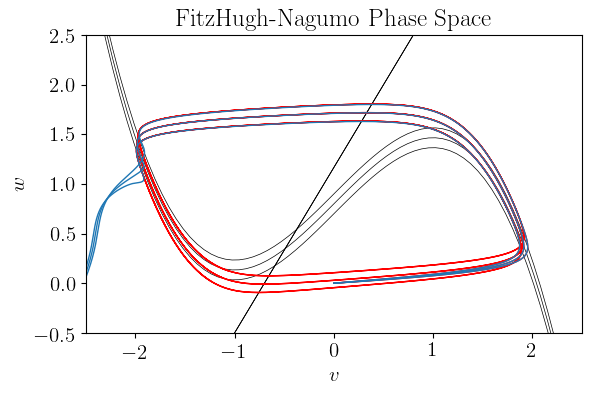}
  \includegraphics[width=0.45\textwidth]{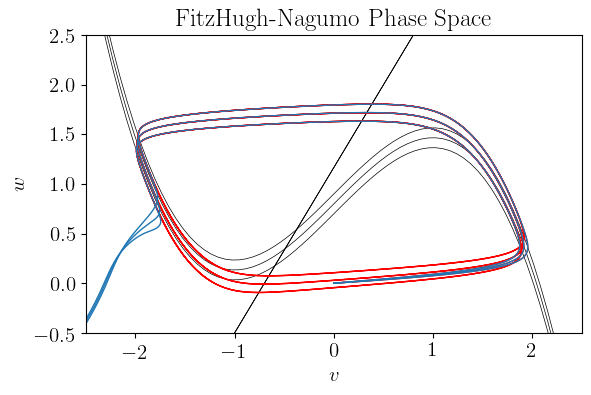}
  \includegraphics[width=0.45\textwidth]{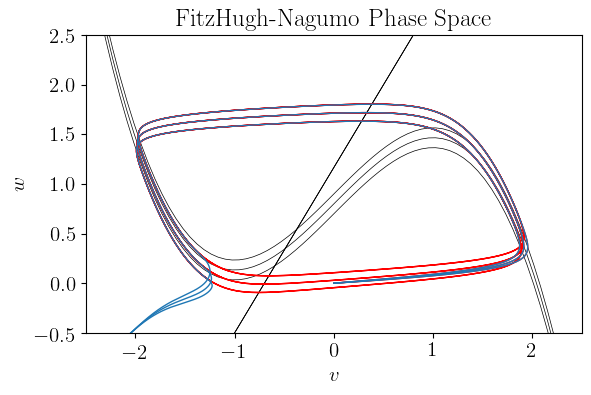}
  \includegraphics[width=0.45\textwidth]{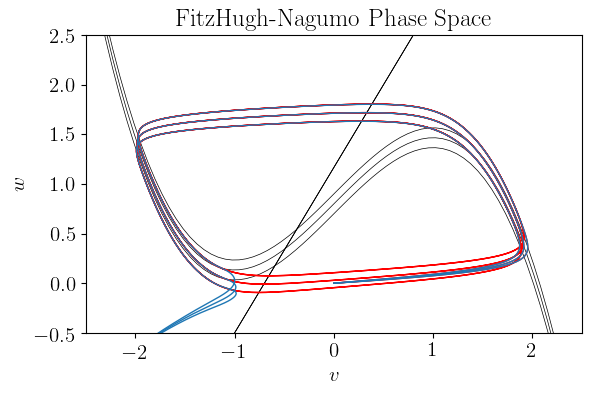}
  \includegraphics[width=0.45\textwidth]{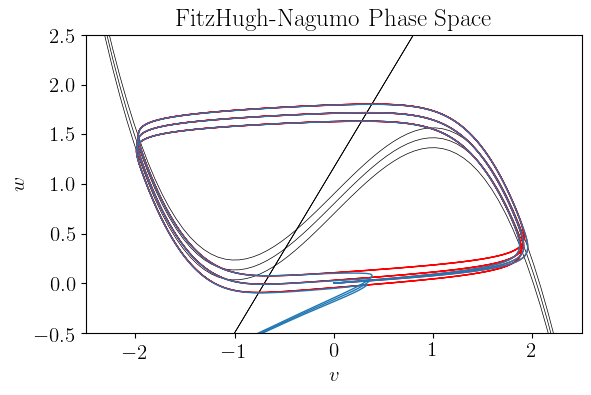}
  \caption{Training FitzHugh-Nagumo solution bundles with curriculum learning. Unlike Figure \ref{fig:fhnbadappendix}, the nullclines are safely passed.}
  \label{fig:fhngoodappendix}
\end{figure}

\clearpage
\subsection{Efficiency Tests: Simple Harmonic Oscillator}

\textbf{ODE:}
\begin{align}
  \dv{x}{t} = v, \qquad \dv{v}{t} = -\frac{k}{m} x,
\end{align}
where the position of the harmonic oscillator is given by $x$, its velocity is $v$. The parameters of the system, spring constant $k$ and mass $m$, are both fixed at unity.

\begin{table}[h]
  \begin{tabular}{llll}
    \toprule
    Network & \multicolumn{3}{l}{Architecture} \\
    \midrule
  1 & Input layer & 3 neurons & $\tanh$ \\
  & Hidden layers & 2 dense layers of 4 neurons each & $\tanh$ \\
  & Output layer & 2 neurons & linear\\
    \midrule
  2 & Input layer & 3 neurons & $\tanh$ \\
  & Hidden layers & 2 dense layers of 8 neurons each & $\tanh$ \\
  & Output layer & 2 neurons & linear\\
    \midrule
  3 & Input layer & 3 neurons & $\tanh$ \\
  & Hidden layers & 2 dense layers of 16 neurons each & $\tanh$ \\
  & Output layer & 2 neurons & linear\\
    \midrule
  4 & Input layer & 3 neurons & $\tanh$ \\
  & Hidden layers & 2 dense layers of 32 neurons each & $\tanh$ \\
  & Output layer & 2 neurons & linear\\
 \bottomrule
 \end{tabular}
 \caption{Simple harmonic oscillator network architectures.}
\end{table}

\begin{table}[h]
  \begin{tabular}{llll}
    \toprule
    Network & Quantity & Value \\
    \midrule
1 & Weighting function & $b(t) = 1$  & \\
& Initial condition ranges & \multicolumn{2}{l}{$(x,v) \in [-1.0,1.0]\times[-1.0, 1.0]$}  \\
& ODE parameter ranges & \multicolumn{2}{l}{$k \in [0.5,2.0]$}  \\
&Time range & $[-0.01, 2\pi]$ & \\
&Optimizer & Adam & \\
&Batch size & 100,000 & \\
&Learning rate & batch 0 to 4,999,999 & $\eta=0.0001$ \\
&Learning rate & batch 5,000,000 to 9,999,999 & $\eta=0.00002$ \\
&Training rate & 179 batches/sec & \\
&Training time & 15.5 hours & \\
\midrule
2 & Weighting function & $b(t) = 1$  & \\
& Initial condition ranges & \multicolumn{2}{l}{$(x,v) \in [-1.0,1.0]\times[-1.0, 1.0]$}  \\
& ODE parameter ranges & \multicolumn{2}{l}{$k \in [0.5,2.0]$}  \\
&Time range & $[-0.01, 2\pi]$ & \\
&Optimizer & Adam & \\
&Batch size & 100,000 & \\
&Learning rate & batch 0 to 10,999,999 & $\eta=0.0001$ \\
&Learning rate & batch 11,000,000 to 15,999,999 & $\eta=0.00002$ \\
&Training rate & 168 batches/sec & \\
&Training time & 26.5 hours & \\
\midrule
3 & Weighting function & $b(t) = 1$  & \\
& Initial condition ranges & \multicolumn{2}{l}{$(x,v) \in [-1.0,1.0]\times[-1.0, 1.0]$}  \\
& ODE parameter ranges & \multicolumn{2}{l}{$k \in [0.5,2.0]$}  \\
&Time range & $[-0.01, 2\pi]$ & \\
&Optimizer & Adam & \\
&Batch size & 100,000 & \\
&Learning rate & batch 0 to 16,999,999 & $\eta=0.00005$ \\
&Learning rate & batch 17,000,000 to 20,999,999 & $\eta=0.00002$ \\
&Learning rate & batch 21,000,000 to 24,999,999 & $\eta=0.00001$ \\
&Training rate & 152 batches/sec & \\
&Training time & 45.7 hours & \\
\midrule
4 & Weighting function & $b(t) = 1$  & \\
& Initial condition ranges & \multicolumn{2}{l}{$(x,v) \in [-1.0,1.0]\times[-1.0, 1.0]$}  \\
& ODE parameter ranges & \multicolumn{2}{l}{$k \in [0.5,2.0]$}  \\
&Time range & $[-0.01, 2\pi]$ & \\
&Optimizer & Adam & \\
&Batch size & 100,000 & \\
&Learning rate & batch 0 to 19,999,999 & $\eta=0.0001$ \\
&Learning rate & batch 20,000,000 to 24,999,999 & $\eta=0.00002$ \\
&Learning rate & batch 25,000,000 to 29,999,999 & $\eta=0.00001$ \\
&Learning rate & batch 30,000,000 to 44,999,999 & $\eta=0.000002$ \\
&Training rate & 126 batches/sec & \\
&Training time & 99.2 hours & \\
 \bottomrule
 \end{tabular}
 \caption{Simple harmonic oscillator networks' training hyperparameters and other details.}
\end{table}

\begin{figure}[h]
  \centering 
  \includegraphics[width=0.8\textwidth]{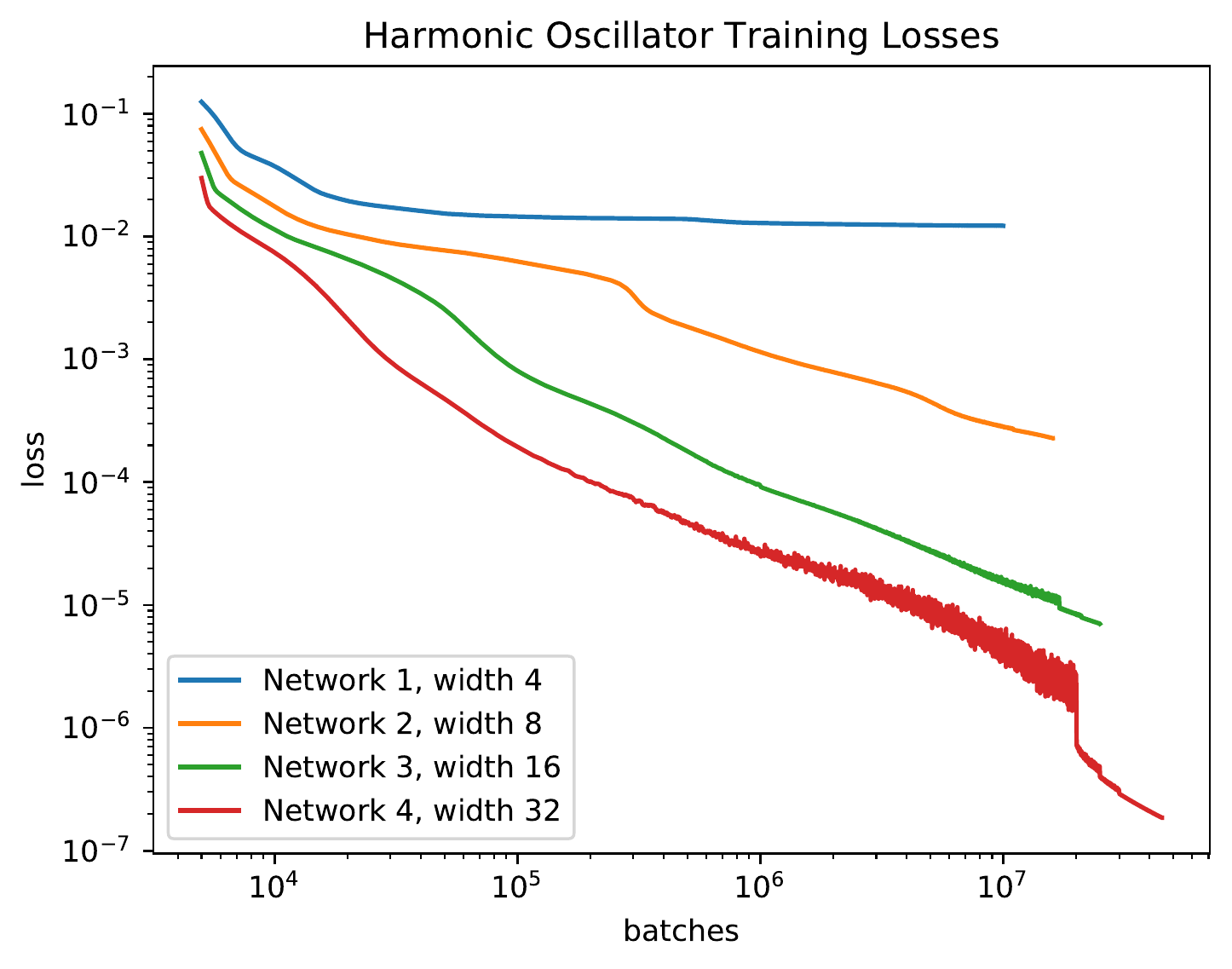}
  \caption{Training losses of the four harmonic oscillator solution bundles, smoothed with a moving average with window size 10,000.}
  \label{fig:sholosses}
\end{figure}

\end{appendices}
\end{document}